\documentclass{article}
\usepackage{microtype}
\usepackage{graphicx}
\usepackage{booktabs}
\usepackage{placeins}
\usepackage{amsmath,amssymb}
\usepackage{xcolor}
\usepackage{tikz}
\usetikzlibrary{positioning,arrows.meta,calc,decorations.pathreplacing}
\usepackage{hyperref}

\usepackage[accepted]{mlsys2025}
\makeatletter\renewcommand{\Notice@String}{}\makeatother
\hypersetup{pdfsubject={}}
\usepackage[capitalise]{cleveref}

\renewcommand{\algorithmiccomment}[1]{\hfill$\triangleright$~#1}
\newcommand{\method}{PLoQ}

\mlsystitlerunning{\method{}: A Prompt-Specific Low-Rank Basis for 4-bit Quantization of LLM Activations}

\begin{document}

\twocolumn[
\mlsystitle{\method{}: A Prompt-Specific Low-Rank Basis for 4-bit Quantization of LLM Activations}

\begin{mlsysauthorlist}
\mlsysauthor{Faraz Tahmasebi}{uci}
\mlsysauthor{Michael Pellauer}{nvidia}
\mlsysauthor{Hyoukjun Kwon}{uci}
\mlsysauthor{Sitao Huang}{uci}
\end{mlsysauthorlist}

\mlsysaffiliation{uci}{Department of Computer Science, University of California, Irvine, CA, USA}
\mlsysaffiliation{nvidia}{NVIDIA, USA}

\mlsyscorrespondingauthor{Faraz Tahmasebi}{tahmasef@uci.edu}
\mlsyskeywords{Quantization, Large Language Models}

\vskip 0.3in

\begin{abstract}
Low-bit activation quantization of large language models remains challenging,
as a small number of dominant directions dictate the quantization range.
State-of-the-art methods isolate these directions in a high-precision subspace through transformations calibrated offline and folded into the weights, which fixes the subspace for all prompts, leaving each prompt’s few unique directions unprotected, and rules it out wherever no weight can absorb the transformation. We present \method{}, a prompt-adaptive low-rank decomposition that extracts the dominant activation components before
quantization, quantizes them separately, and restores them within the layer's
GEMM, without modifying the weights. The low-rank component's basis is estimated
from each prompt during prefill and reused throughout decoding. At the same
bit width as state-of-the-art mixed-precision quantization (4-bit weights,
activations and K/V cache with an eighth of the channels at 8 bits, about 4.2
bits per activation on average), \method{} outperforms it on all six Llama and Qwen models from 1B to 70B parameters,
at up to 8k tokens (16k on two of them), recovering up to 46\% of the perplexity that the state of the art still loses to 4-bit activations. On Llama-3.1-8B at 8k tokens
it adds only 4\% to prefill and 8\% to decoding.
\end{abstract}
]

\printAffiliationsAndNotice{}

\section{Introduction}
\label{sec:intro}

Quantization is the most widely used way to reduce the inference cost of large
language models. Weight-only quantization is mature: 4-bit weights lose
little accuracy~\cite{frantar2023gptq,awq2024}, and vector quantization with
incoherence processing reaches 2 to 3 bits~\cite{quip2023,quipsharp2024,aqlm2024,qtip2024}.
The K/V cache can likewise be stored in 2 to 4 bits~\cite{kivi2024,kvquant2024}.
These methods reduce memory footprint and traffic, but the matrix
multiplications still run in 16-bit arithmetic. Quantizing the activations as
well (W4A4) enables 4-bit tensor-core arithmetic and further lowers memory
traffic~\cite{ashkboos2024quarot,lin2025qserve}.

Activations, however, are challenging to quantize at low precision. A small
number of dominant directions carry magnitudes far exceeding the rest and
dictate the per-token quantization range, leaving the remaining components
with too few effective levels~\cite{dettmers2022llmint8,xiao2023smoothquant}.
Rotation-based methods multiply the activations by orthogonal matrices folded
into the adjacent weights, which spreads the outliers across channels and
makes the distribution smoother~\cite{ashkboos2024quarot,liu2025spinquant,duquant2024,flatquant2025,ostquant2025};
this made W4A4 inference with a 4-bit K/V cache (W4A4KV4) practical.
Mixed-precision methods go further~\cite{zhao2024atom,quik2024,saxena2025resq}: they
identify the dominant channels or directions on calibration data and keep them in a small
8-bit block, an eighth of the channels at three of the four inputs of each
layer, which raises the average activation width from 4 to about 4.2 bits;
they achieve the highest accuracy in this regime. Even so, activation quantization remains the main source of accuracy
loss: on Llama-3.2-1B and Llama-3.1-8B, quantizing the activations adds about
twice the perplexity that quantizing the weights and the K/V cache does
(\cref{tab:main}).

\begin{figure}[t]
\centering
\includegraphics[width=\columnwidth]{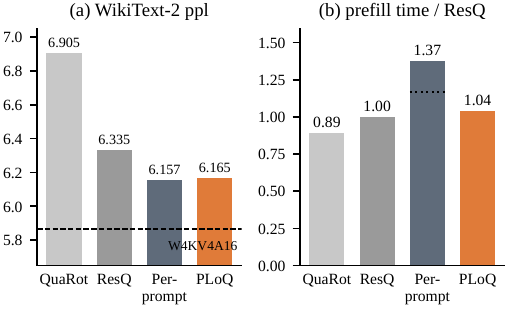}
\caption{Four ways to treat the dominant activation directions, Llama-3.1-8B,
W4A4KV4, 8k tokens. QuaRot~\cite{ashkboos2024quarot}: a rotation, all
activations at 4 bits. ResQ~\cite{saxena2025resq}: a calibrated
eighth of the directions at 8 bits and the rest at 4. Per-prompt: each
prompt's own leading eighth at 8 bits instead, mapped into the 8-bit
channels by a per-prompt block reflection. \method{}: a calibrated block of
224 directions plus a per-prompt rank-32 component with 4-bit coefficients. All but QuaRot
at equal bits. (a) WikiText-2 perplexity. (b) Prefill time relative to ResQ
on an RTX~4090, batch 1. Per-prompt is unfused: ResQ's time plus its
measured extra work; dotted: lower bound, at the GPU's peak fp16 rate.}
\label{fig:motivcmp}
\end{figure}

These transformations, however, are computed offline and folded into the
weights, which imposes two limitations. First, the protected subspace is
static, whereas the dominant directions vary considerably across prompts
(\cref{sec:motivation}), and adapting the whole subspace to each prompt is
costly (\cref{fig:motivcmp}). Second, the transformation is applicable only where
a weight can absorb it. The input of the down projection, the elementwise
product of the gate and up projections in the feed-forward block, has no
weight that can absorb a rotation: rotation-based methods apply only an online
Hadamard transform there, and mixed-precision methods that protect a subspace quantize it
entirely at 4 bits, although its activations are as concentrated in a few
directions as those of the other layer inputs.

We present \method{}, a prompt-adaptive low-rank decomposition for activation
quantization. \method{} does not transform the activation. It \emph{decomposes} it:
the components along a small set of dominant directions are extracted before
quantization, quantized on a dedicated scale, and restored as additional
channels of the layer's GEMM. This changes what can be protected, and when.
Because no basis is folded into the weights, the decomposition applies at every
layer input, including the down-projection input that no offline method can
protect. Because nothing is precomputed, the directions can be chosen per
prompt: \method{} estimates them from each prompt during prefill, with a
randomized range finder fused into the existing kernels, and reuses them
throughout decoding. The low-rank component takes the place of part of the
8-bit block, so the average bit width is unchanged, about 4.2 bits per
activation.

Across Llama and Qwen models from 1B to 70B parameters, \method{} consistently
improves on the strongest mixed-precision method, ResQ~\cite{saxena2025resq}, at equal bit width
and outperforms rotation-based methods on every model. The improvement holds
from 2k to 16k tokens and carries over to zero-shot benchmarks. The prompt-adaptive basis is central to these gains: on
the Llama models, a calibrated basis of the same rank achieves only 55--77\%
of its improvement, and the share is smallest on the largest model
(WikiText-2 perplexity on Llama-3.1-70B: 4.024 with the calibrated 8-bit block
alone, 3.878 with a calibrated low-rank component, 3.758 with the
prompt-adaptive one). On an RTX~4090,
\method{} adds 4\% to prefill latency at 8k tokens.

\noindent Our contributions are as follows:
\begin{itemize}
  \item We show that offline transformations leave a significant share of the
        4-bit activation error unaddressed, as their subspace is static and
        unavailable at some layer inputs, and that capturing a few prompt-specific
        directions significantly reduces the remaining error (\cref{sec:motivation}).
  \item We propose \method{}, a prompt-adaptive low-rank decomposition that
        protects the dominant directions of any layer input at equal bit
        width without modifying the weights: the directions are estimated
        from each prompt and their coefficients restored inside the GEMM
        (\cref{sec:method}).
  \item We make the per-prompt fit practical on GPUs: a single persistent
        kernel fused into the existing producers and overlapped with the
        8-bit GEMM, so that \method{} adds 4\% to prefill at 8k tokens
        (\cref{sec:system}).
  \item We evaluate \method{} on six models from 1B to 70B parameters against
        mixed-precision, rotation-based and outlier-channel methods, and
        analyze the choice of basis, including calibrated and shrinkage
        variants (\cref{sec:eval}).
\end{itemize}

\section{Background and Motivation}
\label{sec:background}

\begin{figure*}[t]
\centering
\resizebox{\textwidth}{!}{\usetikzlibrary{positioning,arrows.meta,calc,decorations.pathreplacing}
\begingroup
\input{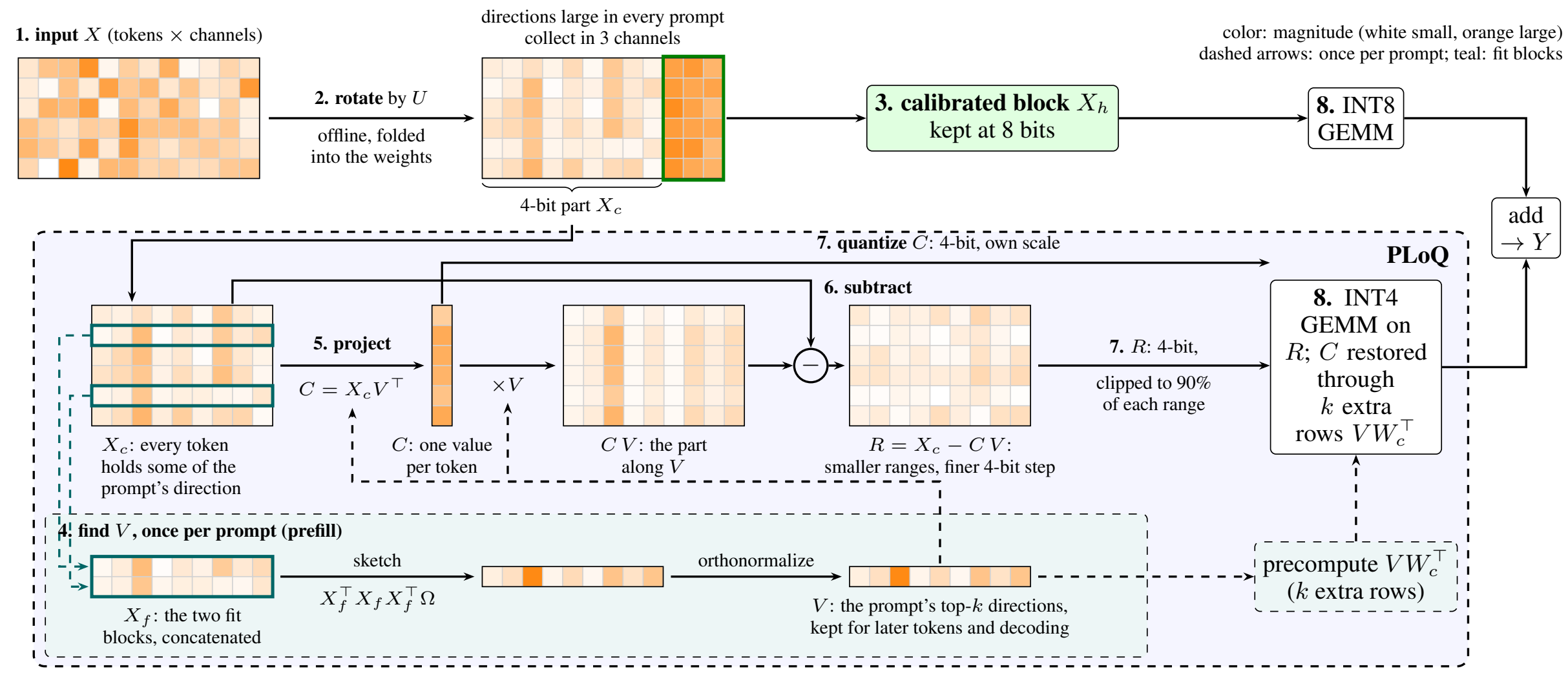}
\begin{tikzpicture}[font=\footnotesize,
  box/.style={draw, rounded corners=2pt, align=center, inner sep=3pt, fill=white},
  g8/.style={box, fill=green!12},
  pre/.style={box, dashed, fill=teal!7},
  arr/.style={-{Stealth[length=4pt]}, thick}, darr/.style={arr, dashed},
  lab/.style={font=\scriptsize, align=center},
  st/.style={font=\scriptsize\bfseries, align=center},
  fitr/.style={draw=teal!80!black, very thick},
  br/.style={decorate, decoration={brace, amplitude=3pt}}]
\def\h{6*\gc}
\def\xz{6.0}
\def\ym{-3*\gc}
\def\xa{0.95}
\def\yb{-3.2}
\def\yf{-6.45}
\filldraw[fill=blue!4, draw=black, dashed, rounded corners=4pt, thick]
  (0.2,\yb+0.95) rectangle (18.75,\yf-1.42);
\filldraw[fill=teal!7, draw=black, dashed, rounded corners=2pt]
  (\xa-0.6,\yf+0.55) rectangle (14.6,\yf-1.3);
\gridX
\node[st, above] at (6*\gc,0.05) {1. input $X$ \textmd{(tokens $\times$ channels)}};
\draw[arr] (12*\gc+0.12,\ym) -- node[lab, above]{\textbf{2. rotate} by $U$} node[lab, below]{offline, folded\\into the weights} (\xz-0.1,\ym);
\begin{scope}[shift={(\xz,0)}] \gridZ
  \draw[very thick, green!50!black] (9*\gc,0.02) rectangle (12*\gc,-\h-0.02); \end{scope}
\node[lab, above] at (\xz+6*\gc,0.05) {directions large in every prompt\\collect in 3 channels};
\draw[br] (\xz+9*\gc-0.02,-\h-0.06) -- node[lab, below=2pt] (xcl) {4-bit part $X_c$} (\xz,-\h-0.06);
\node[g8] (blk) at (12.6,\ym) {\textbf{3. calibrated block} $X_h$\\kept at 8 bits};
\draw[arr] (\xz+12*\gc+0.04,\ym) -- (blk);
\node[box] (g8m) at (17.3,\ym) {\textbf{8.} INT8\\GEMM};
\draw[arr] (blk) -- (g8m);
\node[box] (sum) at (19.5,-2.2) {add\\$\to Y$};
\draw[arr] (g8m) -| (sum);
\begin{scope}[shift={(\xa,\yb)}] \gridXc
  \foreach \r in {1,4} \draw[fitr] (0,-\r*\gc) rectangle (9*\gc,-\r*\gc-\gc); \end{scope}
\draw[arr] (xcl.south) -- ++(0,-0.2) -| (\xa+2*\gc,\yb+0.04);
\node[lab, anchor=north west, align=left] at (\xa,\yb-\h-0.03) {$X_c$: every token\\holds some of the\\prompt's direction};
\def\xc{5.35}
\draw[arr] (\xa+9*\gc+0.1,\yb-3*\gc) -- coordinate[pos=0.5] (pj) (\xc-0.1,\yb-3*\gc);
\node[st, above=1pt] at (pj) {5. project};
\node[lab, below=1pt] (pjl) at (pj) {$C = X_c V^{\top}$};
\begin{scope}[shift={(\xc,\yb)}] \gridC \end{scope}
\node[lab, anchor=north] at (\xc+0.5*\gc,\yb-\h-0.03) {$C$: one value\\per token};
\def\xcv{7.05}
\draw[arr] (\xc+\gc+0.1,\yb-3*\gc) -- coordinate[pos=0.5] (mv) (\xcv-0.1,\yb-3*\gc);
\node[lab, below=1pt] (mvl) at (mv) {$\times V$};
\begin{scope}[shift={(\xcv,\yb)}] \gridCV \end{scope}
\node[lab, anchor=north] at (\xcv+4.5*\gc,\yb-\h-0.03) {$C\,V$: the part\\along $V$};
\node[draw, circle, inner sep=1pt, thick] (minus) at (10.25,\yb-3*\gc) {$-$};
\draw[arr] (\xcv+9*\gc+0.05,\yb-3*\gc) -- (minus);
\draw[arr] (\xa+7*\gc,\yb+0.02) -- ++(0,0.3) -| (minus.north);
\node[st, anchor=south west] at (10.3,\yb+0.03) {6. subtract};
\def\xr{10.75}
\draw[arr] (minus) -- (\xr-0.06,\yb-3*\gc);
\begin{scope}[shift={(\xr,\yb)}] \gridR \end{scope}
\node[lab, anchor=north] at (\xr+4.5*\gc,\yb-\h-0.03) {$R = X_c - C\,V$:\\smaller ranges, finer 4-bit step};
\node[box, minimum height=2.25cm, text width=2.0cm] (g4m) at (17.3,\yb-0.8) {\textbf{8.} INT4 GEMM on $R$; $C$ restored through $k$ extra rows $V W_c^{\top}$};
\draw[arr] (\xc+0.5*\gc,\yb+0.02) -- (\xc+0.5*\gc,\yb+0.55) -- node[lab, above, pos=0.6]{\textbf{7. quantize} $C$: 4-bit, own scale} (\xc+0.5*\gc,\yb+0.55 -| g4m.west);
\draw[arr] (\xr+9*\gc+0.1,\yb-3*\gc) -- node[lab, above]{\textbf{7.} $R$: 4-bit,} node[lab, below]{clipped to 90\%\\of each range} (g4m.west |- 0,\yb-3*\gc);
\draw[arr] (g4m) -| (sum);
\begin{scope}[shift={(\xa,\yf)}] \gridXf \draw[fitr] (0,0) rectangle (9*\gc,-2*\gc); \end{scope}
\node[lab, anchor=north] at (\xa+4.5*\gc,\yf-2*\gc-0.03) {$X_f$: the two fit\\blocks, concatenated};
\draw[darr, teal!80!black] (\xa-0.02,\yb-1.5*\gc) -- (\xa-0.42,\yb-1.5*\gc) |- (\xa-0.04,\yf-0.5*\gc);
\draw[darr, teal!80!black] (\xa-0.02,\yb-4.5*\gc) -- (\xa-0.28,\yb-4.5*\gc) |- (\xa-0.04,\yf-1.5*\gc);
\draw[arr] (\xa+9*\gc+0.1,\yf-\gc) -- node[lab, above]{sketch} node[lab, below]{$X_f^{\top} X_f X_f^{\top}\Omega$} (5.9,\yf-\gc);
\begin{scope}[shift={(6.0,\yf-0.5*\gc)}] \gridYsk \end{scope}
\draw[arr] (6.0+9*\gc+0.1,\yf-\gc) -- node[lab, above]{orthonormalize} (10.65,\yf-\gc);
\begin{scope}[shift={(10.75,\yf-0.5*\gc)}] \gridV \end{scope}
\node[lab, anchor=north] at (10.75+4.5*\gc,\yf-1.5*\gc-0.03) {$V$: the prompt's top-$k$ directions,\\kept for later tokens and decoding};
\draw[darr] (10.75+4.5*\gc,\yf-0.5*\gc+0.04) -- (10.75+4.5*\gc,\yb-\h-0.72) -| (mvl.south);
\draw[darr] (10.75+4.5*\gc,\yb-\h-0.72) -| (pjl.south);
\node[pre] (vw) at (17.3,\yf-\gc) {precompute $V W_c^{\top}$\\($k$ extra rows)};
\draw[darr] (10.75+9*\gc+0.1,\yf-\gc) -- (vw);
\draw[darr] (vw) -- (g4m);
\node[st, anchor=north west] at (\xa-0.6,\yf+0.55) {\,4. find $V$, once per prompt (prefill)};
\node[font=\small\bfseries, anchor=north east] at (18.63,\yb+0.89) {PLoQ};
\node[lab, anchor=north east, align=right] at (20.1,0.55) {color: magnitude (white small, orange large)\\dashed arrows: once per prompt; teal: fit blocks};
\end{tikzpicture}
\endgroup}
\caption{\method{} at one layer input, on an activation example (6 tokens, 12
channels; color: magnitude). Steps 1--3 and the INT8 GEMM are the offline
mixed-precision method: a calibrated basis $U$, folded into the weights,
gathers the directions that are large in every prompt into a block kept at
8 bits. \method{} (dashed box) handles the remaining 4-bit part $X_c$, whose
dominant direction depends on the prompt. Once per prompt, during prefill, it
estimates that direction $V$ from a sample of the prompt's tokens (4). Each
token's component along it, $C = X_c V^{\top}$, is subtracted (5, 6); the
remainder $R$ has a smaller range and is quantized to 4 bits with clipping,
and $C$ to 4 bits on its own scale (7). $C$ is restored through $k$ extra rows
$V W_c^{\top}$ of the 4-bit GEMM (8). $V$ and $V W_c^{\top}$ are kept for
decoding.}
\label{fig:workflow}
\end{figure*}

\subsection{Quantizing activations}

A linear layer computes $Y = XW^\top$ for activations
$X \in \mathbb{R}^{S\times H}$ (one row per token) and weights
$W \in \mathbb{R}^{N\times H}$. In W4A4 inference both operands are quantized
to 4-bit integers. Activations are quantized per token, with one scale per row
set by the row's range, so that the scale can be computed on the fly as each
token arrives. Every entry of a token is then rounded on a grid of step
$\Delta_t = (\max_j x_{tj} - \min_j x_{tj})/15$. LLM activations are
heavy-tailed: a few feature dimensions are consistently much larger than the
rest~\cite{dettmers2022llmint8}, and some tokens produce extreme values in
otherwise quiet channels~\cite{xiao2023smoothquant}. Any such value stretches
its token's $\Delta_t$, and every other entry of that token is rounded
coarsely.

\subsection{Rotating and protecting the dominant directions}
\label{sec:background:offline}

For any orthogonal matrix $Q$, $XW^\top = (XQ)(WQ)^\top$. Transforming the
activations by $Q$ and folding $Q$ into the weights offline therefore changes
the distribution that is quantized without changing the layer's output.
Rotation-based methods choose $Q$ to spread every direction over all
channels: a randomized Hadamard transform makes each channel a mixture of all
the others, so an outlier's energy is shared rather than concentrated, and the
per-token range shrinks~\cite{ashkboos2024quarot}. The rotation can also be
learned, combined with permutations, or extended to affine
transforms~\cite{liu2025spinquant,duquant2024,flatquant2025,ostquant2025}.
All channels are then quantized at the same precision.

Mixed-precision methods instead keep the components that dominate the error
at a higher precision and quantize the rest at 4 bits. Most select
\emph{channels} by a saliency metric. LLM.int8() detects, at run time, the
feature dimensions whose magnitude exceeds a threshold and computes them in
16 bits~\cite{dettmers2022llmint8}; OliVe pairs each outlier with a
neighboring value that it sacrifices, so that outliers can be encoded locally
in hardware~\cite{olive2023}. Atom and QuIK keep a fixed set of outlier
channels, identified on calibration data, at 8 or 16
bits~\cite{zhao2024atom,quik2024}; Tender splits the channels into groups of
similar magnitude with power-of-two related scales~\cite{tender2024}; and
MicroMix gives each channel a 4-, 6- or 8-bit microscaling format according to
its magnitude~\cite{micromix2026}. For weights, OWQ, SpQR and MixLLM select
columns, individual weights or output features by their sensitivity to the
loss~\cite{owq2024,spqr2024,mixllm2026}. A subspace generalizes a channel
selection~\cite{saxena2025resq}: a basis $U$ is computed from the eigenvectors
of the activation covariance on calibration data, and its top-$r$ directions,
which carry most of the variance, form a block kept at 8 bits. The other
$H-r$ directions are rotated among themselves to flatten them and quantized
at 4 bits. With $r = H/8$, the average precision is 4.5 bits per element, and
the two blocks are quantized with separate scales, so the dominant directions
no longer set the scale of the 4-bit part. We call such a block, of channels
or of directions, a \emph{calibrated block}. In all of these methods it is
chosen once, offline, for every prompt. \method{} keeps a calibrated block and
decomposes what it leaves at 4 bits; it applies to any of them, and we
evaluate it with the subspace form, the most accurate at equal bits.

\begin{figure}[t]
\centering
\includegraphics[width=\columnwidth]{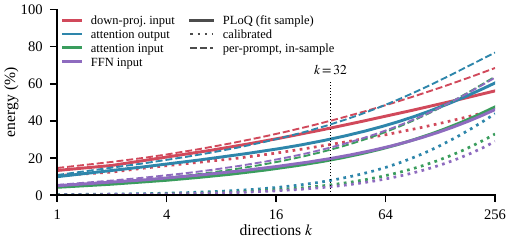}
\caption{Energy of each input's 4-bit part captured by $k$ directions on
held-out tokens: \method{}'s, fitted on the prompt's fit sample (solid); the
calibrated ones (dotted); and the best for the scored tokens themselves
(dashed, in-sample, not deployable). Vertical line: the rank \method{} uses, which reaches 84–98\% of the largest gap to the calibrated directions at a small overhead (Fig. 5). Llama-3.2-1B, eight 8k-token WikiText-2
prompts; held-out tokens are outside the fit sample and after the first
quarter; averaged over layers. The 4-bit part is everything outside the
calibrated 8-bit block; at the down-projection input, which has none, the
whole input.}
\label{fig:energy}
\end{figure}

\subsection{Limits of offline subspaces}
\label{sec:motivation}
\label{sec:background:folding}

A subspace is chosen offline because the basis change into it must be folded
into the weights to be free: into the weight that produces the activation
and, transposed, into the weight that consumes it. In a Llama-style decoder
layer this holds for three of the four projection inputs. The
query/key/value and gate/up inputs are produced by an RMSNorm whose scale is
folded into the next weights, and the residual stream is a sum of linear
outputs, so a basis applied to it can be folded into every weight that writes
to or reads from it. The attention-output input is a per-head combination of
value vectors, whose basis can be folded into the value projection. The
down-projection input is different: it is
$\mathrm{SiLU}(x W_g^\top) \odot (x W_u^\top)$, an elementwise product of two
GEMM outputs. The product commutes with a permutation or a scaling of the
channels, so a channel selection can be folded there, but not with a general
rotation, so no weight can absorb the basis of a protected subspace. A dense
basis change at run time would cost an $H_f \times H_f$ product per token,
more than the down projection itself. Rotation-based methods therefore apply
only an online Hadamard transform at this input, at $O(H_f \log H_f)$ per
token, and subspace methods quantize it entirely at 4 bits. This is a limitation of the
offline transformation rather than of the input, as the measurements below
show. They are taken on Llama-3.2-1B, with subspaces calibrated on WikiText-2
training sequences and evaluated on held-out prompts (16 prompts of 2048
tokens and 8 of 8192; \cref{app:motivation}).

The down-projection input is as concentrated as the others: on held-out
prompts, the top eighth of the calibrated directions hold 75\%, 81\% and 60\%
of the energy of the attention input, attention output and FFN input, and
71\% at the down-projection input. Only the first three get an 8-bit block
that holds this energy. At the down-projection input, the dominant
directions are quantized at 4 bits together with everything else and set each
token's quantization step. A low rank captures much of them: the 32 leading
calibrated directions, 0.4\% of its 8192, carry 27\% of the energy of
held-out tokens (\cref{fig:energy}). At the three protected inputs, the 32 leading directions of
what remains \emph{after} the 8-bit block carry at most 8\%, since the
protected block has already taken the dominant ones. The error follows the
same pattern: a rank-32 component at the down-projection input alone
recovers 72\% of the improvement that the same component reaches at all four
inputs together, and without it the other three inputs are worse than the
offline method (\cref{tab:perinput}).

The dominant directions also depend on the prompt. A prompt's own top-$H/8$
subspace holds 83--97\% of its energy, whereas the calibrated subspace of the
same size holds 60--81\%, and the 32 leading directions of two different
prompts overlap by only 1--6\% (\cref{app:motivation}). A fixed basis
therefore protects directions that are right on average but not for a given
prompt. A sample of the prompt is enough to find its own: fitted on a quarter
of the prompt's tokens, 32 directions hold 19--36\% of the 4-bit energy of
the other tokens, against 5--27\% for the 32 calibrated ones and 25--40\% for
the best possible (\cref{fig:energy}). Adapting the entire protected block to each prompt, however, is
costly (\cref{fig:motivcmp}). Keeping each prompt's own leading eighth at
8 bits instead of the calibrated one lowers the perplexity of Llama-3.1-8B at
8k tokens from 6.335 to 6.157, but it needs a rank-$H/8$ fit, projection and
correction at every input. Even at the GPU's peak arithmetic rate this work
would add 17\% to the prefill time, and measured without fusion it adds 38\%.
A rank-32 per-prompt component on top of the calibrated block, as in \method{}
(\cref{sec:method}), reaches 6.165 for 4\%, and it is better on tokens
after the prompt's first quarter (5.955 against 5.987). The leading directions that the
calibrated block misses are few, and a low-rank component that needs no basis
change can capture them at every layer input, including the down projection.

\section{\method{}: Prompt-Adaptive Low-Rank Quantization}
\label{sec:method}

\begin{figure*}[t]
\centering
\includegraphics[width=0.92\textwidth]{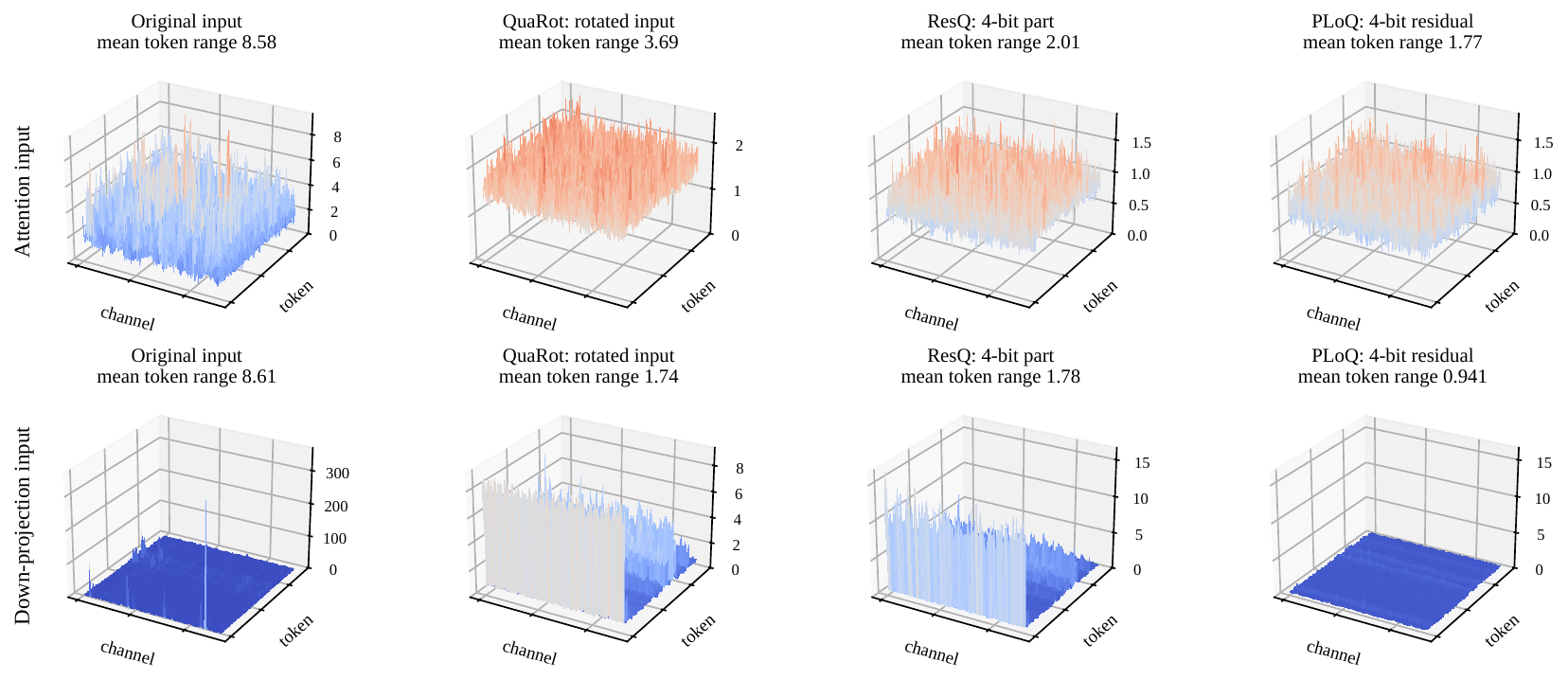}
\caption{Activation magnitudes $|X|$ over tokens and channels (max-pooled to a
$128 \times 128$ grid) for one WikiText-2 prompt on Llama-3.2-1B, layer 15.
From left to right: the original input; after QuaRot's Hadamard rotation,
all at 4 bits; what a calibrated 8-bit block leaves at 4 bits; and what
\method{} leaves at 4 bits after subtracting the prompt's own 32 leading
directions. The last two share one scale. Each title gives the mean per-token
range, which sets the token's 4-bit step. \method{} narrows the calibrated block's range by 12\% at the attention
input and by 47\% at the down-projection input, where a few tokens carry
very large values; averaged over all 16 layers, by 10\% and 27\%.}
\label{fig:activations}
\end{figure*}

The error of 4-bit activation quantization is concentrated in a few dominant
directions of each layer input, each spread across all of its channels
(\cref{sec:motivation}). An offline rotation, folded into the weights at no
runtime cost, maps the calibrated dominant directions onto $r$ coordinates
that are then kept at 8 bits. Directions that depend on the prompt cannot be
protected this way, since mapping them onto coordinates would require
rotating the weights for every prompt, at the cost of the layer itself.
\method{} therefore subtracts these directions instead of rotating them
(\cref{fig:workflow}). At each layer input it measures the activation's
component along $k$ prompt-specific directions, removes it from the part that
the offline rotation leaves at 4 bits, quantizes the $k$ resulting
coefficients on their own scale, and restores their contribution after the
GEMM through a small $k \times N$ matrix. The remainder keeps its width but no
longer contains the dominant directions, so its quantization step is smaller
(\cref{fig:activations}). The calibrated 8-bit block is kept as is; only the
4-bit part of the offline rotation is decomposed.

\subsection{Low-rank decomposition of the 4-bit activations}
\label{sec:method:layer}

Consider a linear layer $Y = XW^\top$ whose input has already been rotated
by the offline basis, so that the first $H-r$ coordinates form the
low-precision part and the last $r$ the protected block. The offline method
quantizes the two separately:
\begin{equation}
Y \approx \mathcal{Q}_4\!\left(X_{\ell}\right) W_{\ell}^\top
        + \mathcal{Q}_8\!\left(X_{h}\right) W_{h}^\top ,
\label{eq:offline}
\end{equation}
where $X_\ell$ and $X_h$ are the two column blocks, $W_\ell, W_h$ the matching
column blocks of $W$, and $\mathcal{Q}_b$ per-token quantization to $b$ bits.
At the down-projection input $r = 0$ (\cref{sec:background:folding}).

\method{} keeps $r' = r-k$ directions in the protected block (none where $r =
0$); the other $c = H - r'$ coordinates form the 4-bit part $X_c$. Given an
orthonormal basis $V \in \mathbb{R}^{k\times c}$ of $k$ directions in it
(\cref{sec:method:basis}):
\begin{align}
C &= X_c V^\top \in \mathbb{R}^{S \times k}, \qquad R = X_c - \mathcal{Q}_4(C)\, V, \\
Y &\approx \mathcal{Q}_4(R)\, W_c^\top + \mathcal{Q}_8(X_{h'})\, W_{h'}^\top
      + \mathcal{Q}_4(C)\, \underbrace{\left(V W_c^\top\right)}_{k \times N} .
\label{eq:method}
\end{align}
Compared with \cref{eq:offline}, only the 4-bit term changes: the 4-bit part
is split into a remainder $R$ and a low-rank component $C V$, whose
contribution is added back after the GEMM. In exact arithmetic \cref{eq:method}
equals $XW^\top$ for any orthonormal $V$; only the quantization error depends
on $V$. No basis is folded into the weights, so the decomposition applies at
every layer input, including the down-projection input, where a dense basis
change would cost an $H_f \times H_f$ product.

Once the dominant directions are removed, the remainder's extreme values are
rare, and clipping its 4-bit range to $\gamma = 90\%$ of the per-token range
improves accuracy at no cost. Only the 4-bit tier is clipped; the 8-bit block
keeps its full range.

\subsection{Choosing the basis}
\label{sec:method:basis}
\label{sec:method:adaptive}

The best $V$ for a prompt spans the top-$k$ right singular subspace of that
prompt's $X_c$: it removes the most energy from the remainder, and with it the
largest per-token ranges. An exact singular value decomposition of every
layer input during prefill would cost far more than the layer. \method{}
instead estimates the subspace with a randomized range
finder~\cite{halko2011finding} on a sample $X_f$ of the prompt's rows, with
one power iteration:
\begin{equation}
Y = X_f^\top X_f X_f^\top \Omega, \qquad
V^\top = Y L^{-\top}, \quad L L^\top = Y^\top Y,
\label{eq:fit}
\end{equation}
where $\Omega \in \mathbb{R}^{S/4 \times k}$ is a fixed random matrix (SRHT in
the accuracy runs, Gaussian in the kernels) and $L$
the Cholesky factor of the $k \times k$ Gram matrix, so that $V$ is
orthonormal (CholeskyQR~\cite{cholqr}). Each additional power iteration
multiplies $Y$ by $X_f^\top X_f$ once more and sharpens the estimate toward
the dominant subspace; one is enough (\cref{sec:system:fit}). Inputs shared by
several projections (query/key/value, gate/up) share one $V$. The basis is
fitted before the prompt is quantized and kept for decoding. Which directions
are removed matters: a random $V$ of the same rank recovers nothing
(\cref{sec:eval:ablation}).

The sample determines both the cost of the fit and which directions it finds.
The dominant directions drift along a prompt, so a sample taken from its
beginning finds the directions of the beginning, and a basis fitted on it
helps the tokens it was fitted on more than the ones that follow. \method{}
therefore samples the whole prompt at the cost of a quarter of it: $X_f$
consists of the rows
\begin{equation}
\mathcal{F} = \{\, t : t \bmod P < b \,\}, \qquad P = 4b,
\label{eq:blocks}
\end{equation}
that is, contiguous blocks of $b$ tokens, one in every $P$, so that each
kernel tile of the fit reads a contiguous range of rows
(\cref{sec:system:fit}). \Cref{tab:fitblocks} compares block sizes at the same
number of fit rows, scored on tokens that no compared fit has seen. Every
spread sample beats the first quarter, on all tokens and on held-out ones, by
up to 0.05 perplexity at 2k and 0.05 at 8k on all tokens, and by up to 0.10
and 0.13 on held-out tokens. Among spread samples the block size matters
little: from $b = 1$ to $b = 128$, perplexity varies by at most 0.03 and
held-out perplexity by at most 0.06, and only the largest block ($b = 256$,
one block per 1024 tokens) begins to lose. We use $b = 64$; a sample of half
the size ($b = 32$, $P = 256$) still reaches 10.058 at 8k.

\begin{table}[t]
\centering\small
\caption{Block size $b$ of the fit sample (\cref{eq:blocks}), a quarter of
the prompt in every row; Llama-3.2-1B, WikiText-2. Held-out: tokens after the
first quarter and outside every compared sample.}
\label{tab:fitblocks}
\setlength{\tabcolsep}{4pt}
\begin{tabular}{lrrrr}
\toprule
 & \multicolumn{2}{c}{2k} & \multicolumn{2}{c}{8k} \\
\cmidrule(lr){2-3}\cmidrule(lr){4-5}
Fit sample & ppl & held-out & ppl & held-out \\
\midrule
First quarter             & 11.357 & 10.587 & 10.074 & 9.925 \\
$b = 256$                 & 11.346 & 10.573 & 10.035 & 9.871 \\
$b = 128$                 & 11.321 & 10.535 & 10.021 & 9.833 \\
$b = 64$ (\method{})        & 11.329 & 10.540 & 10.032 & 9.843 \\
$b = 32$                  & 11.326 & 10.491 & 10.034 & 9.796 \\
$b = 16$                  & 11.305 & 10.503 & 10.039 & 9.856 \\
Every 4th ($b = 1$)       & 11.321 & 10.485 & 10.051 & 9.839 \\
\bottomrule
\end{tabular}
\end{table}

A per-prompt fit is not the only choice. A \emph{calibrated} basis is fixed
offline, as the top-$k$ eigenvectors of the second moment of $X_c$ over
calibration data,
\begin{equation}
\Sigma_{\mathrm{cal}} = \frac{1}{n} \sum_{i=1}^{n} X_{c,i}^\top X_{c,i} ,
\label{eq:cal}
\end{equation}
where $X_{c,i}$ is the 4-bit part of calibration sequence $i$. The fit then
moves offline together with $V W_c^\top$: there is no per-prompt work and no
per-sequence state, at some loss of accuracy (\cref{sec:eval:ablation}).
Between the two, a \emph{shrunk} basis blends the prompt's estimate with the
calibrated one, for models on which a quarter of the prompt is a poor guide to
its directions. $V$ is the top-$k$ eigenspace of
\begin{equation}
(1-\alpha)\,\frac{\Sigma_{\mathrm{cal}}}{\operatorname{tr}\Sigma_{\mathrm{cal}}}
+ \alpha\,\frac{X_f^\top X_f}{\|X_f\|_F^2}, \qquad \alpha = 0.25,
\label{eq:shrink}
\end{equation}
which is the top-$k$ right singular subspace of the stacked matrix
$\bigl[\sqrt{1-\alpha}\,F;\ \sqrt{\alpha}\,X_f / \|X_f\|_F\bigr]$, with
$F$ a rank-512 factor of $\Sigma_{\mathrm{cal}} / \operatorname{tr}
\Sigma_{\mathrm{cal}}$ computed offline. A range finder on this matrix costs
the fit of \cref{eq:fit} plus 512 rows.

\Cref{alg:method} summarizes \method{} at one layer input. Lines 1--5 run once
per prompt during prefill: they fit $V$ and form the $k \times N$ correction
matrix $U = V W_c^\top$. Lines 7--9 run for every token, in prefill and in
decode. \Cref{sec:system} maps each line to a GPU kernel.

\begin{algorithm}[t]
\caption{\method{} at one layer input.}
\label{alg:method}
\begin{algorithmic}[1]
\REQUIRE activations $X = [X_c,\, X_{h'}]$ after the offline rotation
  ($S$ prompt tokens); weights $W_c, W_{h'}$; random
  $\Omega \in \mathbb{R}^{S/4 \times k}$; clip ratio $\gamma$
\STATE $X_f \gets$ rows $\mathcal{F}$ of $X_c$ \COMMENT{spaced blocks, $S/4$ rows}
\STATE $Y_0 \gets X_f^\top \Omega$ \COMMENT{sketch}
\STATE $Z \gets X_f Y_0$;\ \ $Y \gets X_f^\top Z$ \COMMENT{power iteration}
\STATE $V^\top \gets \mathrm{CholeskyQR}(Y)$ \COMMENT{$c \times k$, orthonormal}
\STATE $U \gets V W_c^\top$ \COMMENT{$k$ extra GEMM rows}
\FOR{each prefill block and each decode token $X$}
  \STATE $C \gets X_c V^\top$;\ \ $R \gets X_c - \mathcal{Q}_4(C)\, V$ \COMMENT{producer}
  \STATE $\hat{Y} \gets \mathcal{Q}_4^{\gamma}(R)\, W_c^\top
         + \mathcal{Q}_8(X_{h'})\, W_{h'}^\top$ \COMMENT{GEMM}
  \STATE $\hat{Y} \gets \hat{Y} + \mathcal{Q}_4(C)\, U$ \COMMENT{epilogue}
\ENDFOR
\ENSURE $\hat{Y} \approx X W^\top$; $V$, $U$ kept for decoding
\end{algorithmic}
\end{algorithm}

\subsection{Equal average precision}
\label{sec:method:bits}

Per token and layer input, the offline method stores $8r + 4(H-r)$ bits.
\method{} stores $r-k$ values at 8 bits, $H-r+k$ residual values at 4 bits, and
$k$ coefficients at 4 bits:
\begin{equation}
8(r-k) + 4(H-r+k) + 4k \;=\; 8r + 4(H-r).
\end{equation}
Trading $k$ protected directions for a rank-$k$ component keeps the average
bits per element unchanged. At the down-projection input, where $r = 0$, the
$k$ coefficients are additional: 0.2\% of the model's activation bits for
$k = 32$ on Llama-3.2-1B. The K/V cache is quantized as in the offline method.

\section{Kernel Fusion for \method{}}
\label{sec:system}

\Cref{alg:method} adds four operations to a layer input: the fit (lines 1--4,
once per prompt), forming the correction matrix $U$ (line 5), the projection
and subtraction (line 7), and the correction (line 9). For $k = 32$ their
arithmetic is under 1\% of the layer's GEMMs, so on a GPU their cost is
latency and memory traffic, not computation. A direct implementation, with
each operation as its own kernel, would add several passes over every
activation and one over every output. Our kernels follow one rule instead: add no
pass over a layer's output and as few as possible over its input, by fusing
each operation into a kernel that the baseline already runs or appending it to
an existing GEMM. The baseline, a calibrated 8-bit block with the rest at 4 bits, and
\method{} share one W4A4 pipeline that we implement and tune for both, with
int4 and int8 tensor-core GEMMs in CUTLASS and the producers and epilogues in
Triton. This
section follows a prompt through prefill, where the basis is fitted and $U$
formed, and then through decoding, where both are reused.

\subsection{Fitting the basis during prefill}
\label{sec:system:fit}

The fit's products, $X_f^\top \Omega$, $X_f Y_0$ and $X_f^\top Z$ (lines 2--3), are thin: $k =
32$ columns against a sample of $S/4$ rows, so they are bounded by memory
bandwidth. At the attention and FFN inputs, the activation is the RMSNorm of
the residual stream, $X = \mathrm{diag}(\rho)\, h\, \mathrm{diag}(w)$, so every
product against $X$ is a product against $h$ with the $k$-wide operand scaled
by $\rho$ or $w$: the fit reads the residual stream directly, with each
token's normalization factor $\rho$ taken from the sum of squares that the
preceding residual addition accumulates in its epilogue, and never writes the
normalized sample. At the down-projection input, the sketch of the fit rows
is accumulated inside the SwiGLU kernel that produces them. The kernels map a
tile's rows to the spaced blocks of \cref{eq:blocks}, so the sample is never
gathered into a separate buffer. These producer fusions pay off once a
batch holds about 8k tokens or more; for shorter batches, separate kernels are
faster, and we use them there. The fit is a chain of dependent steps, each too small to fill the GPU: three
thin products, a column normalization after each of the first two, the Gram
matrix of the result, its Cholesky factor, and the orthonormalization.
Launched separately they are about ten kernels per layer input, and their cost
is the latency of the chain rather than its work. \method{} runs the three
products, both normalizations and the partial Gram matrices in one persistent
kernel, whose thread blocks are all resident at once and pass from one step to
the next through a barrier in global memory; the sample is read from memory by
the first product and found in the L2 cache by the other two. A single warp
then factors the $k \times k$ Gram matrix (CholeskyQR~\cite{cholqr}), and a
last small kernel applies the factor. Intermediate orthonormalizations would
not change the span, so a column normalization between the products keeps the
iteration stable at no cost.

Even fused, the fit sits on the critical path: the projection, the quantizer
and the GEMM all wait for $V$. The int8 GEMM of the calibrated block does not:
its input is the block itself, which does not depend on $V$. \method{} therefore
quantizes the calibrated block and runs its int8 GEMM on a second stream while
the fit runs, and the int4 GEMM then accumulates into that GEMM's output
instead of the other way around. The two GEMMs do the same work as before, and
the fit, which occupies few of the GPU's multiprocessors, overlaps with a GEMM
that keeps the memory system busy; at 8k tokens it is hidden entirely at the
attention and FFN inputs. The down-projection input has no 8-bit block, and
its fit remains exposed. For prompts under 8k tokens, the int4 copy of $V$
for the GEMM and the packed basis kept for decoding are also produced on a
side stream, beside the quantizer. The fits of all prompts in a batch are
launched together, so their launch latency is paid once per batch.

The power iteration (line 3) sharpens the sketch toward the dominant
subspace (\cref{tab:poweriter}). The first iteration is essential, and it is free: without
it, perplexity rises from 11.329 to 11.436 at 2k tokens and from 10.032 to
10.148 at 8k, a sixth to a fifth of the gain, while the prefill
time changes by less than a point, since the fit runs beside the 8-bit GEMM. A
second iteration improves perplexity by only 0.004 at 2k and 0.024 at 8k but
adds 4.4 points of prefill overhead at 2k and 1.4 at 8k, and \method{} therefore
uses one.

\begin{table}[t]
\centering\small
\caption{Power iterations: quality (Llama-3.2-1B, WikiText-2; in parentheses,
held-out tokens as in \cref{tab:fitblocks}) and prefill overhead (Llama-3.1-8B shapes, RTX~4090, batch
1). For zero and two iterations, the change in overhead relative to one
iteration, in percentage points, timed in one run in which every row runs
the down-projection fit outside the SwiGLU kernel; for one iteration, the
overhead from \cref{tab:overhead}.}
\label{tab:poweriter}
\resizebox{\columnwidth}{!}{%
\begin{tabular}{lrrrr}
\toprule
 & \multicolumn{2}{c}{2k} & \multicolumn{2}{c}{8k} \\
\cmidrule(lr){2-3}\cmidrule(lr){4-5}
Iterations & ppl & prefill & ppl & prefill \\
\midrule
0           & 11.436 (10.605) & $-$0.4 & 10.148 (9.937) & +0.7 \\
1 (\method{}) & 11.329 (10.540) & +14.2\% & 10.032 (9.843) & +4.1\% \\
2           & 11.325 (10.500) & +4.4 & 10.008 (9.793) & +1.4 \\
\bottomrule
\end{tabular}}
\end{table}

$U = V W_c^\top$ (line 5) would cost a second read of the layer's weights if
computed on its own. Instead, the $k$ rows of $V$ are appended to the
activation operand of the layer's int4 GEMM during prefill, so $U$ comes out
of the same pass over the weights, at the cost of $k$ more rows in a GEMM of
$S$. It is then stored at 8 bits for decoding.
\label{sec:system:u}

\subsection{Projection, subtraction and correction}
\label{sec:system:fuse}

Every token, in prefill and in decode, goes through lines 7--9. The producer
of each layer input, the kernel that quantizes it for the GEMM, also performs
the subtraction: it reads the activation and the coefficients $C$, and writes
the remainder's 4-bit codes, its clipped per-token scale and the quantized
coefficients in one launch. The projection $C = X_c V^\top$ needs every
coordinate of a token before the subtraction can start, so it is computed
where the full token is already in hand: in the RMSNorm pass at the attention
and FFN inputs, and in the SwiGLU kernel at the down-projection input, which
projects onto $V$ before the online Hadamard transform by using the
transformed basis. The correction $\mathcal{Q}_4(C)\,U$ is a rank-$k$ update
of the GEMM output, fused into the epilogue of the kernel that reads that
output anyway: RoPE after the query/key/value projection, the residual
addition after the attention-output and down projections, and SwiGLU after
gate/up. It adds no pass over the output.

The remaining prefill cost is the fit at the down-projection input, the
subtraction inside the quantizer, the $k$ extra GEMM rows and the padding of
the wider 4-bit tier to the GEMM's tile, and a few small kernels per input. On
Llama-3.1-8B at 8k tokens these add 15\,ms to the baseline's 379\,ms
(\cref{sec:eval:overhead}).

\subsection{Decoding}
\label{sec:system:decode}

During decoding, $V$ and $U$ are reused and nothing is fitted. A single kernel
per layer input performs the normalization, the projection, the subtraction
and both quantizer passes on the new token. The projection is a reduction over
the whole token, which would cost the kernel a synchronization across thread
blocks; instead, the epilogue of the kernel that produces the token, the
residual addition or SwiGLU, accumulates the token's sum of squares and its
projection onto the next input's $V$ as it writes the token, and the producer
waits on a single barrier. The sum of squares is offered to the baseline's producers
as well, since they need it for RMSNorm. Each sequence keeps $V$ at 4 bits and
$U$ at 8 bits, 58.5\,MB for Llama-3.1-8B, about the K/V cache of 1.4k tokens;
the down-projection producer also keeps its basis multiplied by the Hadamard
transform in 16 bits, another 29\,MB. With a calibrated basis, lines 1--5 move
offline: $U$ is stored with the weights, and there is no fit and no
per-sequence state.

\section{Evaluation}
\label{sec:eval}

\subsection{Setup}
\label{sec:eval:setup}

\paragraph{Models and quantization.} Llama-3.2-1B and 3B, Llama-3.1-8B and
70B~\cite{llama3}, Qwen2.5-7B and 32B~\cite{qwen25}, all at W4A4KV4. Weights
are quantized with GPTQ~\cite{frantar2023gptq}, activations per token, and the
K/V cache with one eighth of each head at 8 bits. The main baseline is
ResQ~\cite{saxena2025resq}, with $r = H/8$ protected directions at 8 bits
(4.5 bits per element at those inputs, about 4.2 over all inputs). \method{} uses the same basis, rotations and
average bits, with $r - 32$ protected directions and a rank-32 component at
every input. We also run QuaRot~\cite{ashkboos2024quarot} and
QuIK~\cite{quik2024} in the same pipeline. QuIK keeps, at each layer, the
eighth of the channels with the largest mean magnitude on calibration data at
8 bits, at the same three inputs and without a rotation (GPTQ damping 0.1 on
Llama-3.1-70B). We quote
SpinQuant~\cite{liu2025spinquant} as reported by~\citet{saxena2025resq}. Each
method is reported at its better clipping setting, and all use the
open-source pipeline of~\citet{saxena2025resq}. We chose $\gamma$ and $b$ on
the test data.

\paragraph{Metrics.} Perplexity on WikiText-2~\cite{wikitext} at 2k, 8k and
16k tokens; zero-shot accuracy on eight commonsense tasks and
MMLU~\cite{mmlu} with lm-evaluation-harness~\cite{lmeval}, on the first
500 examples of each commonsense task and 25 of each MMLU subject. The
\emph{recovered fraction}
$(\mathrm{ppl}_{\text{ResQ}} - \mathrm{ppl}_{\method{}}) /
 (\mathrm{ppl}_{\text{ResQ}} - \mathrm{ppl}_{\text{W4KV4A16}})$ is the share of
the error that 4-bit activations add which \method{} removes. To check that the
per-prompt basis does not merely fit the tokens it is scored on, we also score
\emph{held-out} tokens: those outside the fit sample and after the first
quarter of the prompt.

\paragraph{Latency.} End to end on an RTX~4090, with CUDA graphs and int4 and
int8 tensor-core GEMMs for both methods. ResQ runs on its fastest kernels, and
the quantization kernels the two methods share are tuned for both
(\cref{sec:system}). \Cref{app:a100} repeats the measurement on an A100.

\subsection{Main results}
\label{sec:eval:main}

\begin{table}[t]
\centering\small
\caption{WikiText-2 perplexity at 2k tokens, W4A4KV4. ResQ and \method{} at
equal average activation bits. Rec.\ = recovered fraction of the 4-bit
activation error. $^\ast$ Reported by~\citet{saxena2025resq}, one decimal;
Llama-3 for the 8B and 70B columns; no W4A4KV4 results are published for
Qwen2.5, which the released code does not support. $^\dagger$ No Hadamard matrix of the FFN
width exists (\citet{saxena2025resq} report 4035.9).
}
\label{tab:main}
\setlength{\tabcolsep}{4pt}
\resizebox{\columnwidth}{!}{%
\begin{tabular}{lrrrrrr}
\toprule
Method & \multicolumn{2}{c}{Llama-3.2} & Llama-3.1 & \multicolumn{2}{c}{Qwen2.5} & Llama-3.1 \\
 & 1B & 3B & 8B & 7B & 32B & 70B \\
\midrule
16-bit           & 9.761  & 7.828 & 6.249 & 6.850 & 5.023 & 2.839 \\
W4KV4A16         & 10.452 & 8.114 & 6.531 & 8.003 & 5.198 & 3.246 \\
\midrule
QuIK             & 16.641 & 10.506 & 8.356 & $2.3{\times}10^4$ & 6.916 & 7.434 \\
SpinQuant$^\ast$ & 13.6   & 9.2   & 7.4   & --    & --    & 6.2   \\
QuaRot           & 14.464 & 9.949 & 7.721 & 3624$^\dagger$ & 5.919 & 5.302 \\
ResQ             & 12.005 & 8.680 & 7.093 & 8.396 & 5.564 & 4.024 \\
\method{}          & \textbf{11.329} & \textbf{8.451} & \textbf{6.835} & \textbf{8.354} & \textbf{5.431} & \textbf{3.758} \\
\midrule
Rec.\ (\method{})  & 43\% & 40\% & 46\% & 11\% & 36\% & 34\% \\
\bottomrule
\end{tabular}}
\end{table}

\begin{table}[t]
\centering\small
\caption{WikiText-2 perplexity at long context, W4A4KV4, as in
\cref{tab:main}. $^\dagger$ QuaRot fails (4266). No long-context results are
published for SpinQuant.}
\label{tab:longctx}
\setlength{\tabcolsep}{4pt}
\resizebox{\columnwidth}{!}{%
\begin{tabular}{lrrrrrrrr}
\toprule
Method & \multicolumn{3}{c}{Llama-3.2} & \multicolumn{2}{c}{Llama-3.1-8B} & \multicolumn{2}{c}{Qwen2.5} & Llama-3.1 \\
 & 1B & 1B & 3B & & & 7B & 32B & 70B \\
 & 8k & 16k & 8k & 8k & 16k & 8k & 8k & 8k \\
\midrule
16-bit   & 8.652 & 8.394 & 6.989 & 5.614 & 5.478 & 6.129 & 4.469 & 2.567 \\
W4KV4A16 & 9.259 & 8.995 & 7.244 & 5.867 & 5.726 & 7.386 & 4.633 & 2.939 \\
\midrule
QuIK     & 14.923 & 14.769 & 9.606 & 7.472 & 7.282 & $1.9{\times}10^4$ & 6.178 & 6.708 \\
QuaRot   & 12.700 & 12.355 & 8.804 & 6.905 & 6.724 & --$^\dagger$ & 5.254 & 4.774 \\
ResQ     & 10.561 & 10.210 & 7.731 & 6.335 & 6.174 & 7.679 & 4.945 & 3.648 \\
\method{}  & \textbf{10.032} & \textbf{9.754} & \textbf{7.552} & \textbf{6.165} & \textbf{6.009} & \textbf{7.566} & \textbf{4.842} & \textbf{3.426} \\
\midrule
Rec.\ (\method{}) & 41\% & 38\% & 37\% & 36\% & 37\% & 39\% & 33\% & 31\% \\
\bottomrule
\end{tabular}}
\end{table}

\paragraph{Perplexity.} At the same average bits, \method{} improves on ResQ on
every model and at every context length (\cref{tab:main,tab:longctx}). On all
models except Qwen2.5-7B it recovers 34--46\% of the perplexity lost to 4-bit
activations at 2k tokens and 31--41\% at 8k. It outperforms QuaRot and QuIK on every model, and the published SpinQuant
numbers. The gain shrinks only slowly with context: on
Llama-3.2-1B, from 43\% at 2k to 41\% at 8k and 38\% at 16k. It also holds on
held-out tokens: on Llama-3.2-1B at 2k, 10.958 for ResQ against 10.406 for
\method{}.

\paragraph{Qwen2.5-7B.} This model gains the least at 2k (11\%) but 39\% at
8k. Clipping helps ResQ much more on Qwen than on Llama, and ResQ is reported
with clipping here. QuaRot fails on this model, as \citet{saxena2025resq} also
report, and so does QuIK (perplexity above $10^4$ with or without clipping).

\begin{table}[t]
\centering\small
\caption{Zero-shot accuracy (\%): mean of eight commonsense tasks, and MMLU.
\method{} fits on the whole (short) prompt. $^\ast$ Reported
by~\citet{saxena2025resq} under their protocol (full test sets, normalized
accuracy on five tasks, 0-shot MMLU; Llama-3 for the 8B and 70B columns; none
published for Qwen2.5),
where their ResQ scores 50.1, 59.0, 63.9 and 71.1, and 29.4, 49.8, 57.2 and
74.0 on MMLU.}
\label{tab:zeroshot}
\setlength{\tabcolsep}{4pt}
\resizebox{\columnwidth}{!}{%
\begin{tabular}{lrrrrrr}
\toprule
Method & \multicolumn{2}{c}{Llama-3.2} & Llama-3.1 & \multicolumn{2}{c}{Qwen2.5} & Llama-3.1 \\
 & 1B & 3B & 8B & 7B & 32B & 70B \\
\midrule
\multicolumn{7}{l}{\emph{Commonsense, mean of eight tasks}} \\
QuIK    & 44.7 & 51.8 & 57.7 & 31.5 & 61.0 & 58.2$^\ast$ \\
QuaRot  & 45.1 & 53.9 & 59.2 & 36.1 & 62.2 & 63.2 \\
ResQ    & 47.9 & 56.0 & 60.7 & 60.6 & 64.3 & 65.5 \\
\method{} & \textbf{48.2} & \textbf{56.8} & \textbf{61.1} & \textbf{62.0} & \textbf{65.0} & \textbf{67.3} \\
\cmidrule(lr){1-7}
SpinQuant$^\ast$ & 48.8 & 57.9 & 63.8 & -- & -- & 65.7 \\
\midrule
\multicolumn{7}{l}{\emph{MMLU}} \\
QuIK    & 28.0 & 39.1 & 54.7 & 23.3 & 74.1 & 51.1$^\ast$ \\
QuaRot  & 27.7 & 43.4 & 54.8 & 24.2 & 78.9 & 71.0 \\
ResQ    & 28.1 & 49.6 & 58.8 & 69.8 & 80.0 & 74.9 \\
\method{} & \textbf{33.7} & \textbf{52.2} & \textbf{61.9} & \textbf{71.1} & \textbf{81.5} & \textbf{75.4} \\
\cmidrule(lr){1-7}
SpinQuant$^\ast$ & 25.6 & 44.2 & 56.2 & -- & -- & 59.4 \\
\bottomrule
\end{tabular}}
\end{table}

\paragraph{Zero-shot accuracy.} The perplexity gains carry over to downstream
tasks (\cref{tab:zeroshot}). \method{} is ahead of ResQ on every model, on both
the commonsense mean and MMLU, and QuaRot is below both throughout. The gains
are 0.5 to 5.6 points on MMLU, largest on the small models, and 0.3 to 1.8 on
the commonsense mean, largest on Llama-3.1-70B.

\subsection{Ablations}
\label{sec:eval:ablation}

\begin{table}[t]
\centering\small
\caption{Building up \method{} on Llama-3.2-1B, WikiText-2, equal bits.}
\label{tab:ablation}
\resizebox{\columnwidth}{!}{%
\begin{tabular}{lrr}
\toprule
Configuration & 2k & 8k \\
\midrule
Calibrated block, 256 directions at 8 bits & 12.005 & 10.561 \\
224 calibrated only & 12.067 & 10.595 \\
\quad + rank-32 per-prompt component & 11.595 & 10.275 \\
\quad + clipping the 4-bit tier & 11.357 & 10.074 \\
\quad + fit on spaced blocks (\method{}) & \textbf{11.329} & \textbf{10.032} \\
\midrule
Random basis instead of per-prompt & -- & 10.599 \\
Calibrated basis instead of per-prompt & 11.485 & 10.112 \\
\bottomrule
\end{tabular}}
\end{table}

\paragraph{Components.} \Cref{tab:ablation} builds up \method{}. Giving up 32
calibrated directions alone costs a little; the per-prompt component more than
recovers it; clipping the 4-bit tier, which only pays once the dominant
directions are gone, and fitting on spaced blocks add the rest. The directions
matter: a random basis of the same rank recovers nothing. Most of the gain
comes from the down-projection input, which the offline method leaves entirely
at 4 bits: on Llama-3.2-1B at 8k, the component there alone gives 72\% of the
improvement for 2.7\% of prefill time (\cref{tab:perinput}).

\paragraph{Per-prompt versus calibrated basis.} A basis calibrated offline
needs no fit, but recovers less (\cref{tab:basismodels}). On the Llama models
at 2k it reaches 55--77\% of the per-prompt basis's improvement, and
the least on the largest: on Llama-3.1-70B it recovers 19\% of the 4-bit
activation error, against 34\%. Shrinking the per-prompt basis toward the
calibrated one lies in between on every Llama model, and the per-prompt basis
stays ahead of the calibrated one on held-out tokens. Qwen2.5-7B is the exception: there the calibrated basis is
best at 8k and the shrunk one at 2k. We measured overhead only for the
per-prompt basis.

\begin{figure}[t]
\centering
\includegraphics[width=\columnwidth]{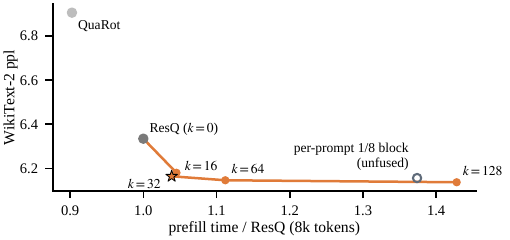}
\caption{Per-prompt rank $k$ at equal bits: $512-k$ calibrated directions and
$k$ per-prompt ones ($k=0$ is ResQ), Llama-3.1-8B, WikiText-2 at 8k tokens.
Prefill time relative to ResQ (RTX~4090, batch 1); kernels are tuned for
$k=32$ (star) only. Hollow: whole 8-bit block per prompt (\cref{fig:motivcmp}).}
\label{fig:ksweep}
\end{figure}

\paragraph{How many per-prompt directions?} \Cref{fig:ksweep} trades
calibrated directions for per-prompt ones at equal bits. The first 16 give
most of the gain (6.335 to 6.181), and 32 reach 6.165 at the same prefill
time. $k=64$ and 128 reach 6.147 and 6.138, below adapting the whole 8-bit
block (6.157), but add 11\% and 43\% to prefill, against 4\% for $k=32$,
which we use.

\subsection{Overhead}
\label{sec:eval:overhead}

\begin{table}[t]
\centering\small
\caption{Overhead of \method{} over ResQ on its fastest kernels, Llama-3.1-8B
shapes, RTX~4090, W4A4 kernels for both, batch 1; median of three runs. Decode:
one step after a prefill of the given length.}
\label{tab:overhead}
\begin{tabular}{lrr}
\toprule
 & 2k & 8k \\
\midrule
Prefill & +14.2\% & +4.1\% \\
Prefill, fit on 1/8 of the prompt & +14.3\% & +2.4\% \\
Decode, per step & +8.1\% & +7.9\% \\
\bottomrule
\end{tabular}
\end{table}

\Cref{tab:overhead} gives the cost of \method{}: the fit, the projection and
correction at each input, the extra GEMM rows, and the per-sequence state.
Prefill overhead falls with prompt length, from 14\% at 2k to 4\% at 8k
tokens: at 8k the fit is hidden behind the 8-bit GEMM at three of the four
layer inputs (\cref{sec:system:fit}), and the fixed cost of each input's extra
kernels is amortized over more tokens. Fitting on 1/8 of the prompt instead of
1/4 lowers the 8k overhead to 2.4\%, for 0.02--0.03 perplexity on
Llama-3.2-1B; at 2k it saves nothing, since the fixed costs dominate. Decode costs 0.6\,ms more per step, and the per-sequence state (58.5\,MB for Llama-3.1-8B) is
about the K/V cache of 1.4k tokens; with the down-projection basis kept in
16 bits for decoding (\cref{sec:system:decode}), 87.5\,MB.

\begin{table}[t]
\centering\small
\caption{Against FP16: Llama-3.1-8B shapes, RTX~4090, CUDA graphs, same
attention kernel and K/V cache layout; in parentheses, speedup over FP16.
FP16 uses cuBLAS GEMMs with the elementwise operations fused. Weights: all
parameters, with the embedding and LM head in 16 bits for all three. Decode
at 8k: one step after a prefill of 8k tokens. Batch 8: eight prompts of 2k
tokens.}
\label{tab:fp16}
\setlength{\tabcolsep}{3pt}
\resizebox{\columnwidth}{!}{%
\begin{tabular}{lrrr}
\toprule
 & FP16 & ResQ & \method{} \\
\midrule
Weights (GB)                 & 16.06 & 5.92 (2.7$\times$) & 5.91 (2.7$\times$) \\
Per-sequence state, 8k (MB)  & 1074  & 336 & 424 \\
\midrule
Prefill, 2k (ms)             & 193.2 & 68.7 (2.81$\times$)  & 78.4 (2.46$\times$) \\
Prefill, 8k (ms)             & 836.2 & 379.3 (2.20$\times$) & 394.7 (2.12$\times$) \\
Decode at 8k (ms/step)       & 17.66 & 7.86 (2.25$\times$)  & 8.47 (2.08$\times$) \\
\midrule
Prefill, batch 8 (k tokens/s) & 10.7 & 27.2 (2.53$\times$) & 24.3 (2.26$\times$) \\
Decode, batch 8 (tokens/s)    & 419  & 952 (2.27$\times$)  & 820 (1.95$\times$) \\
\bottomrule
\end{tabular}}
\end{table}

\paragraph{Against FP16.} \Cref{tab:fp16} puts both W4A4 methods next to
the 16-bit model. \method{} needs no weights of its own: its 8-bit block is
slightly narrower and the coefficients need no stored weights, so the model is
2.7$\times$ smaller than in FP16, as ResQ's. It is $2.1$--$2.5\times$ faster
than FP16 in prefill and $2.1\times$ in decoding at batch 1, against
ResQ's $2.2$--$2.8\times$ and $2.25\times$. Per sequence, it adds its basis
and correction rows to the 4-bit K/V cache, 26\% more than ResQ's at 8k
tokens and still 2.5$\times$ less than FP16's. With eight sequences, decoding
pays the per-sequence correction eight times while the weight GEMMs are
shared, and \method{}'s decode overhead over ResQ rises from 8\% to 16\%.

\section{Related Work}
\label{sec:related}

\paragraph{Quantizing LLMs.} Weight-only methods reach 4 bits and below with
second-order rounding or activation-aware scaling~\cite{frantar2023gptq,awq2024}.
Quantizing activations as well is harder because of outlier
channels~\cite{dettmers2022llmint8}. SmoothQuant migrates their range into the
weights~\cite{xiao2023smoothquant}, OmniQuant learns the clipping and scaling
parameters~\cite{omniquant2024}, and quantization-aware training recovers the
rest at a training cost~\cite{llmqat2024}. Systems work makes W4A4 and W4A8
fast on GPUs~\cite{zhao2024atom,quik2024,lin2025qserve}. K/V caches are often
quantized per channel for keys~\cite{kivi2024,kvquant2024}.

\paragraph{Rotations.} QuaRot applies randomized Hadamard transforms that are
folded into the weights, plus online Hadamard transforms where folding is not
possible~\cite{ashkboos2024quarot}. SpinQuant learns the
rotations~\cite{liu2025spinquant}; DuQuant combines rotations with a
permutation that spreads outliers across blocks~\cite{duquant2024}; FlatQuant
and OSTQuant learn affine or orthogonal-plus-scaling transforms per
layer~\cite{flatquant2025,ostquant2025}. All are fixed before
deployment; at the down-projection input, where no weight can absorb them,
they are applied online and keep the whole input at one precision.

\paragraph{Mixed precision.} \Cref{sec:background:offline} reviews the
methods that keep salient channels or a calibrated subspace at a higher
precision~\cite{dettmers2022llmint8,olive2023,zhao2024atom,quik2024,tender2024,micromix2026,saxena2025resq}.
Their selection is made offline, or, in LLM.int8(), at run time but per
channel. \method{} keeps whatever block they protect and adds a per-prompt
low-rank component at every layer input.

\paragraph{Low-rank corrections.} Low-rank branches are used to compensate
\emph{weight} quantization error: LoRC and LQER add a low-rank approximation of
the weight error~\cite{zeroquantv2,lqer2024}, and SVDQuant absorbs outliers
into a 16-bit low-rank weight branch for W4A4 diffusion
models~\cite{svdquant2025}. These branches correct weight error in 16 bits;
\method{}'s component removes dominant activation directions before
quantization and carries them at 4 bits, within the same bit budget.

\paragraph{Input-dependent quantization.} Rotated runtime smoothing computes smoothing factors from the
activations at run time~\cite{rrs2024}, and PrefixQuant isolates outlier tokens
in a prefixed K/V cache~\cite{prefixquant2024}. \method{} instead
fits the dominant directions of each layer input to the prompt during prefill
and keeps them for decoding.

\section{Conclusion}
\label{sec:conclusion}

Offline transformations protect the dominant directions of LLM activations
only through a static subspace, and only where a weight can
absorb the basis. \method{} lifts both restrictions: it extracts each prompt's
dominant low-rank components before quantization, quantizes them separately,
and restores them in the layer's GEMM, at every layer input at equal
bit width. On six models from 1B to 70B, it improves on state-of-the-art mixed-precision quantization and
outperforms rotation-based methods at modest overhead. The gain holds as
sequences grow (16k where measured), and its edge over a calibrated basis
tends to grow with model size.

\bibliographystyle{mlsys2025}
\bibliography{references}

\begin{thebibliography}{39}
\providecommand{\natexlab}[1]{#1}
\providecommand{\url}[1]{\texttt{#1}}
\expandafter\ifx\csname urlstyle\endcsname\relax
  \providecommand{\doi}[1]{doi: #1}\else
  \providecommand{\doi}{doi: \begingroup \urlstyle{rm}\Url}\fi

\bibitem[Ashkboos et~al.(2024{\natexlab{a}})Ashkboos, Markov, Frantar, Zhong,
  Wang, Ren, Hoefler, and Alistarh]{quik2024}
Ashkboos, S., Markov, I., Frantar, E., Zhong, T., Wang, X., Ren, J., Hoefler,
  T., and Alistarh, D.
\newblock {QUIK}: Towards end-to-end 4-bit inference on generative large
  language models.
\newblock In \emph{Proceedings of the 2024 Conference on Empirical Methods in
  Natural Language Processing (EMNLP)}, pp.\  3355--3371, 2024{\natexlab{a}}.
\newblock \doi{10.18653/v1/2024.emnlp-main.197}.

\bibitem[Ashkboos et~al.(2024{\natexlab{b}})Ashkboos, Mohtashami, Croci, Li,
  Cameron, Jaggi, Alistarh, Hoefler, and Hensman]{ashkboos2024quarot}
Ashkboos, S., Mohtashami, A., Croci, M.~L., Li, B., Cameron, P., Jaggi, M.,
  Alistarh, D., Hoefler, T., and Hensman, J.
\newblock {QuaRot}: Outlier-free 4-bit inference in rotated {LLMs}.
\newblock In \emph{Advances in Neural Information Processing Systems
  (NeurIPS)}, 2024{\natexlab{b}}.

\bibitem[Chee et~al.(2023)Chee, Cai, Kuleshov, and De~Sa]{quip2023}
Chee, J., Cai, Y., Kuleshov, V., and De~Sa, C.
\newblock {QuIP}: 2-bit quantization of large language models with guarantees.
\newblock In \emph{Advances in Neural Information Processing Systems
  (NeurIPS)}, 2023.

\bibitem[Chen et~al.(2026)Chen, Liu, Wang, Bin, Shao, and Luo]{prefixquant2024}
Chen, M., Liu, Y., Wang, J., Bin, Y., Shao, W., and Luo, P.
\newblock {PrefixQuant}: Eliminating outliers by prefixed tokens for large
  language models quantization.
\newblock \emph{IEEE Transactions on Pattern Analysis and Machine
  Intelligence}, 48\penalty0 (11):\penalty0 13972--13983, 2026.
\newblock \doi{10.1109/TPAMI.2026.3711802}.

\bibitem[Dettmers et~al.(2022)Dettmers, Lewis, Belkada, and
  Zettlemoyer]{dettmers2022llmint8}
Dettmers, T., Lewis, M., Belkada, Y., and Zettlemoyer, L.
\newblock {LLM.int8()}: 8-bit matrix multiplication for transformers at scale.
\newblock In \emph{Advances in Neural Information Processing Systems
  (NeurIPS)}, 2022.

\bibitem[Dettmers et~al.(2024)Dettmers, Svirschevski, Egiazarian, Kuznedelev,
  Frantar, Ashkboos, Borzunov, Hoefler, and Alistarh]{spqr2024}
Dettmers, T., Svirschevski, R., Egiazarian, V., Kuznedelev, D., Frantar, E.,
  Ashkboos, S., Borzunov, A., Hoefler, T., and Alistarh, D.
\newblock {SpQR}: A sparse-quantized representation for near-lossless {LLM}
  weight compression.
\newblock In \emph{International Conference on Learning Representations
  (ICLR)}, 2024.

\bibitem[Egiazarian et~al.(2024)Egiazarian, Panferov, Kuznedelev, Frantar,
  Babenko, and Alistarh]{aqlm2024}
Egiazarian, V., Panferov, A., Kuznedelev, D., Frantar, E., Babenko, A., and
  Alistarh, D.
\newblock Extreme compression of large language models via additive
  quantization.
\newblock In \emph{Proceedings of the 41st International Conference on Machine
  Learning (ICML)}, 2024.

\bibitem[Frantar et~al.(2023)Frantar, Ashkboos, Hoefler, and
  Alistarh]{frantar2023gptq}
Frantar, E., Ashkboos, S., Hoefler, T., and Alistarh, D.
\newblock {GPTQ}: Accurate post-training quantization for generative
  pre-trained transformers.
\newblock In \emph{International Conference on Learning Representations
  (ICLR)}, 2023.

\bibitem[Gao et~al.(2023)Gao, Tow, Abbasi, Biderman, Black, DiPofi, Foster,
  Golding, Hsu, Le~Noac'h, Li, McDonell, Muennighoff, Ociepa, Phang, Reynolds,
  Schoelkopf, Skowron, Sutawika, Tang, Thite, Wang, Wang, and Zou]{lmeval}
Gao, L., Tow, J., Abbasi, B., Biderman, S., Black, S., DiPofi, A., Foster, C.,
  Golding, L., Hsu, J., Le~Noac'h, A., Li, H., McDonell, K., Muennighoff, N.,
  Ociepa, C., Phang, J., Reynolds, L., Schoelkopf, H., Skowron, A., Sutawika,
  L., Tang, E., Thite, A., Wang, B., Wang, K., and Zou, A.
\newblock A framework for few-shot language model evaluation, December 2023.
\newblock URL \url{https://zenodo.org/records/10256836}.

\bibitem[Grattafiori et~al.(2024)Grattafiori, Dubey, Jauhri, Pandey, Kadian,
  Al-Dahle, Letman, Mathur, Schelten, Vaughan, Yang, Fan, Goyal, Hartshorn,
  Yang, Mitra, Sravankumar, Korenev, Hinsvark, Rao, Zhang, Rodriguez,
  Gregerson, Spataru, Roziere, Biron, Tang, Chern, Caucheteux, Nayak, Bi,
  Marra, McConnell, Keller, Touret, Wu, Wong, Ferrer, Nikolaidis, Allonsius,
  Song, Pintz, Livshits, Wyatt, Esiobu, Choudhary, Mahajan, Garcia-Olano,
  Perino, Hupkes, Lakomkin, AlBadawy, Lobanova, Dinan, Smith, Radenovic,
  Guzmán, Zhang, Synnaeve, Lee, Anderson, Thattai, Nail, Mialon, Pang,
  Cucurell, Nguyen, Korevaar, Xu, Touvron, Zarov, Ibarra, Kloumann, Misra,
  Evtimov, Zhang, Copet, Lee, Geffert, Vranes, Park, Mahadeokar, Shah, van~der
  Linde, Billock, Hong, Lee, Fu, Chi, Huang, Liu, Wang, Yu, Bitton, Spisak,
  Park, Rocca, Johnstun, Saxe, Jia, Alwala, Prasad, Upasani, Plawiak, Li,
  Heafield, Stone, El-Arini, Iyer, Malik, Chiu, Bhalla, Lakhotia,
  Rantala-Yeary, van~der Maaten, Chen, Tan, Jenkins, Martin, Madaan, Malo,
  Blecher, Landzaat, de~Oliveira, Muzzi, Pasupuleti, Singh, Paluri, Kardas,
  Tsimpoukelli, Oldham, Rita, Pavlova, Kambadur, Lewis, Si, Singh, Hassan,
  Goyal, Torabi, Bashlykov, Bogoychev, Chatterji, Zhang, Duchenne, Çelebi,
  Alrassy, Zhang, Li, Vasic, Weng, Bhargava, Dubal, Krishnan, Koura, Xu, He,
  Dong, Srinivasan, Ganapathy, Calderer, Cabral, Stojnic, Raileanu, Maheswari,
  Girdhar, Patel, Sauvestre, Polidoro, Sumbaly, Taylor, Silva, Hou, Wang,
  Hosseini, Chennabasappa, Singh, Bell, Kim, Edunov, Nie, Narang, Raparthy,
  Shen, Wan, Bhosale, Zhang, Vandenhende, Batra, Whitman, Sootla, Collot,
  Gururangan, Borodinsky, Herman, Fowler, Sheasha, Georgiou, Scialom,
  Speckbacher, Mihaylov, Xiao, Karn, Goswami, Gupta, Ramanathan, Kerkez,
  Gonguet, Do, Vogeti, Albiero, Petrovic, Chu, Xiong, Fu, Meers, Martinet,
  Wang, Wang, Tan, Xia, Xie, Jia, Wang, Goldschlag, Gaur, Babaei, Wen, Song,
  Zhang, Li, Mao, Coudert, Yan, Chen, Papakipos, Singh, Srivastava, Jain,
  Kelsey, Shajnfeld, Gangidi, Victoria, Goldstand, Menon, Sharma, Boesenberg,
  Baevski, Feinstein, Kallet, Sangani, Teo, Yunus, Lupu, Alvarado, Caples, Gu,
  Ho, Poulton, Ryan, Ramchandani, Dong, Franco, Goyal, Saraf, Chowdhury,
  Gabriel, Bharambe, Eisenman, Yazdan, James, Maurer, Leonhardi, Huang, Loyd,
  Paola, Paranjape, Liu, Wu, Ni, Hancock, Wasti, Spence, Stojkovic, Gamido,
  Montalvo, Parker, Burton, Mejia, Liu, Wang, Kim, Zhou, Hu, Chu, Cai, Tindal,
  Feichtenhofer, Gao, Civin, Beaty, Kreymer, Li, Adkins, Xu, Testuggine, David,
  Parikh, Liskovich, Foss, Wang, Le, Holland, Dowling, Jamil, Montgomery,
  Presani, Hahn, Wood, Le, Brinkman, Arcaute, Dunbar, Smothers, Sun, Kreuk,
  Tian, Kokkinos, Ozgenel, Caggioni, Kanayet, Seide, Florez, Schwarz, Badeer,
  Swee, Halpern, Herman, Sizov, Zhang, Lakshminarayanan, Inan, Shojanazeri,
  Zou, Wang, Zha, Habeeb, Rudolph, Suk, Aspegren, Goldman, Zhan, Damlaj,
  Molybog, Tufanov, Leontiadis, Veliche, Gat, Weissman, Geboski, Kohli, Lam,
  Asher, Gaya, Marcus, Tang, Chan, Zhen, Reizenstein, Teboul, Zhong, Jin, Yang,
  Cummings, Carvill, Shepard, McPhie, Torres, Ginsburg, Wang, Wu, U, Saxena,
  Khandelwal, Zand, Matosich, Veeraraghavan, Michelena, Li, Jagadeesh, Huang,
  Chawla, Huang, Chen, Garg, A, Silva, Bell, Zhang, Guo, Yu, Moshkovich,
  Wehrstedt, Khabsa, Avalani, Bhatt, Mankus, Hasson, Lennie, Reso, Groshev,
  Naumov, Lathi, Keneally, Liu, Seltzer, Valko, Restrepo, Patel, Vyatskov,
  Samvelyan, Clark, Macey, Wang, Hermoso, Metanat, Rastegari, Bansal,
  Santhanam, Parks, White, Bawa, Singhal, Egebo, Usunier, Mehta, Laptev, Dong,
  Cheng, Chernoguz, Hart, Salpekar, Kalinli, Kent, Parekh, Saab, Balaji,
  Rittner, Bontrager, Roux, Dollar, Zvyagina, Ratanchandani, Yuvraj, Liang,
  Alao, Rodriguez, Ayub, Murthy, Nayani, Mitra, Parthasarathy, Li, Hogan,
  Battey, Wang, Howes, Rinott, Mehta, Siby, Bondu, Datta, Chugh, Hunt, Dhillon,
  Sidorov, Pan, Mahajan, Verma, Yamamoto, Ramaswamy, Lindsay, Lindsay, Feng,
  Lin, Zha, Patil, Shankar, Zhang, Zhang, Wang, Agarwal, Sajuyigbe, Chintala,
  Max, Chen, Kehoe, Satterfield, Govindaprasad, Gupta, Deng, Cho, Virk,
  Subramanian, Choudhury, Goldman, Remez, Glaser, Best, Koehler, Robinson, Li,
  Zhang, Matthews, Chou, Shaked, Vontimitta, Ajayi, Montanez, Mohan, Kumar,
  Mangla, Ionescu, Poenaru, Mihailescu, Ivanov, Li, Wang, Jiang, Bouaziz,
  Constable, Tang, Wu, Wang, Wu, Gao, Kleinman, Chen, Hu, Jia, Qi, Li, Zhang,
  Zhang, Adi, Nam, Wang, Zhao, Hao, Qian, Li, He, Rait, DeVito, Rosnbrick, Wen,
  Yang, Zhao, and Ma]{llama3}
Grattafiori, A., Dubey, A., Jauhri, A., Pandey, A., Kadian, A., Al-Dahle, A.,
  Letman, A., Mathur, A., Schelten, A., Vaughan, A., Yang, A., Fan, A., Goyal,
  A., Hartshorn, A., Yang, A., Mitra, A., Sravankumar, A., Korenev, A.,
  Hinsvark, A., Rao, A., Zhang, A., Rodriguez, A., Gregerson, A., Spataru, A.,
  Roziere, B., Biron, B., Tang, B., Chern, B., Caucheteux, C., Nayak, C., Bi,
  C., Marra, C., McConnell, C., Keller, C., Touret, C., Wu, C., Wong, C.,
  Ferrer, C.~C., Nikolaidis, C., Allonsius, D., Song, D., Pintz, D., Livshits,
  D., Wyatt, D., Esiobu, D., Choudhary, D., Mahajan, D., Garcia-Olano, D.,
  Perino, D., Hupkes, D., Lakomkin, E., AlBadawy, E., Lobanova, E., Dinan, E.,
  Smith, E.~M., Radenovic, F., Guzmán, F., Zhang, F., Synnaeve, G., Lee, G.,
  Anderson, G.~L., Thattai, G., Nail, G., Mialon, G., Pang, G., Cucurell, G.,
  Nguyen, H., Korevaar, H., Xu, H., Touvron, H., Zarov, I., Ibarra, I.~A.,
  Kloumann, I., Misra, I., Evtimov, I., Zhang, J., Copet, J., Lee, J., Geffert,
  J., Vranes, J., Park, J., Mahadeokar, J., Shah, J., van~der Linde, J.,
  Billock, J., Hong, J., Lee, J., Fu, J., Chi, J., Huang, J., Liu, J., Wang,
  J., Yu, J., Bitton, J., Spisak, J., Park, J., Rocca, J., Johnstun, J., Saxe,
  J., Jia, J., Alwala, K.~V., Prasad, K., Upasani, K., Plawiak, K., Li, K.,
  Heafield, K., Stone, K., El-Arini, K., Iyer, K., Malik, K., Chiu, K., Bhalla,
  K., Lakhotia, K., Rantala-Yeary, L., van~der Maaten, L., Chen, L., Tan, L.,
  Jenkins, L., Martin, L., Madaan, L., Malo, L., Blecher, L., Landzaat, L.,
  de~Oliveira, L., Muzzi, M., Pasupuleti, M., Singh, M., Paluri, M., Kardas,
  M., Tsimpoukelli, M., Oldham, M., Rita, M., Pavlova, M., Kambadur, M., Lewis,
  M., Si, M., Singh, M.~K., Hassan, M., Goyal, N., Torabi, N., Bashlykov, N.,
  Bogoychev, N., Chatterji, N., Zhang, N., Duchenne, O., Çelebi, O., Alrassy,
  P., Zhang, P., Li, P., Vasic, P., Weng, P., Bhargava, P., Dubal, P.,
  Krishnan, P., Koura, P.~S., Xu, P., He, Q., Dong, Q., Srinivasan, R.,
  Ganapathy, R., Calderer, R., Cabral, R.~S., Stojnic, R., Raileanu, R.,
  Maheswari, R., Girdhar, R., Patel, R., Sauvestre, R., Polidoro, R., Sumbaly,
  R., Taylor, R., Silva, R., Hou, R., Wang, R., Hosseini, S., Chennabasappa,
  S., Singh, S., Bell, S., Kim, S.~S., Edunov, S., Nie, S., Narang, S.,
  Raparthy, S., Shen, S., Wan, S., Bhosale, S., Zhang, S., Vandenhende, S.,
  Batra, S., Whitman, S., Sootla, S., Collot, S., Gururangan, S., Borodinsky,
  S., Herman, T., Fowler, T., Sheasha, T., Georgiou, T., Scialom, T.,
  Speckbacher, T., Mihaylov, T., Xiao, T., Karn, U., Goswami, V., Gupta, V.,
  Ramanathan, V., Kerkez, V., Gonguet, V., Do, V., Vogeti, V., Albiero, V.,
  Petrovic, V., Chu, W., Xiong, W., Fu, W., Meers, W., Martinet, X., Wang, X.,
  Wang, X., Tan, X.~E., Xia, X., Xie, X., Jia, X., Wang, X., Goldschlag, Y.,
  Gaur, Y., Babaei, Y., Wen, Y., Song, Y., Zhang, Y., Li, Y., Mao, Y., Coudert,
  Z.~D., Yan, Z., Chen, Z., Papakipos, Z., Singh, A., Srivastava, A., Jain, A.,
  Kelsey, A., Shajnfeld, A., Gangidi, A., Victoria, A., Goldstand, A., Menon,
  A., Sharma, A., Boesenberg, A., Baevski, A., Feinstein, A., Kallet, A.,
  Sangani, A., Teo, A., Yunus, A., Lupu, A., Alvarado, A., Caples, A., Gu, A.,
  Ho, A., Poulton, A., Ryan, A., Ramchandani, A., Dong, A., Franco, A., Goyal,
  A., Saraf, A., Chowdhury, A., Gabriel, A., Bharambe, A., Eisenman, A.,
  Yazdan, A., James, B., Maurer, B., Leonhardi, B., Huang, B., Loyd, B., Paola,
  B.~D., Paranjape, B., Liu, B., Wu, B., Ni, B., Hancock, B., Wasti, B.,
  Spence, B., Stojkovic, B., Gamido, B., Montalvo, B., Parker, C., Burton, C.,
  Mejia, C., Liu, C., Wang, C., Kim, C., Zhou, C., Hu, C., Chu, C.-H., Cai, C.,
  Tindal, C., Feichtenhofer, C., Gao, C., Civin, D., Beaty, D., Kreymer, D.,
  Li, D., Adkins, D., Xu, D., Testuggine, D., David, D., Parikh, D., Liskovich,
  D., Foss, D., Wang, D., Le, D., Holland, D., Dowling, E., Jamil, E.,
  Montgomery, E., Presani, E., Hahn, E., Wood, E., Le, E.-T., Brinkman, E.,
  Arcaute, E., Dunbar, E., Smothers, E., Sun, F., Kreuk, F., Tian, F.,
  Kokkinos, F., Ozgenel, F., Caggioni, F., Kanayet, F., Seide, F., Florez,
  G.~M., Schwarz, G., Badeer, G., Swee, G., Halpern, G., Herman, G., Sizov, G.,
  Zhang, G., Lakshminarayanan, G., Inan, H., Shojanazeri, H., Zou, H., Wang,
  H., Zha, H., Habeeb, H., Rudolph, H., Suk, H., Aspegren, H., Goldman, H.,
  Zhan, H., Damlaj, I., Molybog, I., Tufanov, I., Leontiadis, I., Veliche,
  I.-E., Gat, I., Weissman, J., Geboski, J., Kohli, J., Lam, J., Asher, J.,
  Gaya, J.-B., Marcus, J., Tang, J., Chan, J., Zhen, J., Reizenstein, J.,
  Teboul, J., Zhong, J., Jin, J., Yang, J., Cummings, J., Carvill, J., Shepard,
  J., McPhie, J., Torres, J., Ginsburg, J., Wang, J., Wu, K., U, K.~H., Saxena,
  K., Khandelwal, K., Zand, K., Matosich, K., Veeraraghavan, K., Michelena, K.,
  Li, K., Jagadeesh, K., Huang, K., Chawla, K., Huang, K., Chen, L., Garg, L.,
  A, L., Silva, L., Bell, L., Zhang, L., Guo, L., Yu, L., Moshkovich, L.,
  Wehrstedt, L., Khabsa, M., Avalani, M., Bhatt, M., Mankus, M., Hasson, M.,
  Lennie, M., Reso, M., Groshev, M., Naumov, M., Lathi, M., Keneally, M., Liu,
  M., Seltzer, M.~L., Valko, M., Restrepo, M., Patel, M., Vyatskov, M.,
  Samvelyan, M., Clark, M., Macey, M., Wang, M., Hermoso, M.~J., Metanat, M.,
  Rastegari, M., Bansal, M., Santhanam, N., Parks, N., White, N., Bawa, N.,
  Singhal, N., Egebo, N., Usunier, N., Mehta, N., Laptev, N.~P., Dong, N.,
  Cheng, N., Chernoguz, O., Hart, O., Salpekar, O., Kalinli, O., Kent, P.,
  Parekh, P., Saab, P., Balaji, P., Rittner, P., Bontrager, P., Roux, P.,
  Dollar, P., Zvyagina, P., Ratanchandani, P., Yuvraj, P., Liang, Q., Alao, R.,
  Rodriguez, R., Ayub, R., Murthy, R., Nayani, R., Mitra, R., Parthasarathy,
  R., Li, R., Hogan, R., Battey, R., Wang, R., Howes, R., Rinott, R., Mehta,
  S., Siby, S., Bondu, S.~J., Datta, S., Chugh, S., Hunt, S., Dhillon, S.,
  Sidorov, S., Pan, S., Mahajan, S., Verma, S., Yamamoto, S., Ramaswamy, S.,
  Lindsay, S., Lindsay, S., Feng, S., Lin, S., Zha, S.~C., Patil, S., Shankar,
  S., Zhang, S., Zhang, S., Wang, S., Agarwal, S., Sajuyigbe, S., Chintala, S.,
  Max, S., Chen, S., Kehoe, S., Satterfield, S., Govindaprasad, S., Gupta, S.,
  Deng, S., Cho, S., Virk, S., Subramanian, S., Choudhury, S., Goldman, S.,
  Remez, T., Glaser, T., Best, T., Koehler, T., Robinson, T., Li, T., Zhang,
  T., Matthews, T., Chou, T., Shaked, T., Vontimitta, V., Ajayi, V., Montanez,
  V., Mohan, V., Kumar, V.~S., Mangla, V., Ionescu, V., Poenaru, V.,
  Mihailescu, V.~T., Ivanov, V., Li, W., Wang, W., Jiang, W., Bouaziz, W.,
  Constable, W., Tang, X., Wu, X., Wang, X., Wu, X., Gao, X., Kleinman, Y.,
  Chen, Y., Hu, Y., Jia, Y., Qi, Y., Li, Y., Zhang, Y., Zhang, Y., Adi, Y.,
  Nam, Y., Wang, Y., Zhao, Y., Hao, Y., Qian, Y., Li, Y., He, Y., Rait, Z.,
  DeVito, Z., Rosnbrick, Z., Wen, Z., Yang, Z., Zhao, Z., and Ma, Z.
\newblock The {Llama} 3 herd of models.
\newblock \emph{arXiv preprint arXiv:2407.21783}, 2024.

\bibitem[Guo et~al.(2023)Guo, Tang, Hu, Leng, Zhang, Yang, Liu, Guo, and
  Zhu]{olive2023}
Guo, C., Tang, J., Hu, W., Leng, J., Zhang, C., Yang, F., Liu, Y., Guo, M., and
  Zhu, Y.
\newblock {OliVe}: Accelerating large language models via hardware-friendly
  outlier-victim pair quantization.
\newblock In \emph{Proceedings of the 50th Annual International Symposium on
  Computer Architecture (ISCA)}, 2023.
\newblock \doi{10.1145/3579371.3589038}.

\bibitem[Halko et~al.(2011)Halko, Martinsson, and Tropp]{halko2011finding}
Halko, N., Martinsson, P.-G., and Tropp, J.~A.
\newblock Finding structure with randomness: Probabilistic algorithms for
  constructing approximate matrix decompositions.
\newblock \emph{SIAM Review}, 53\penalty0 (2):\penalty0 217--288, 2011.

\bibitem[Hendrycks et~al.(2021)Hendrycks, Burns, Basart, Zou, Mazeika, Song,
  and Steinhardt]{mmlu}
Hendrycks, D., Burns, C., Basart, S., Zou, A., Mazeika, M., Song, D., and
  Steinhardt, J.
\newblock Measuring massive multitask language understanding.
\newblock In \emph{International Conference on Learning Representations
  (ICLR)}, 2021.

\bibitem[Hooper et~al.(2024)Hooper, Kim, Mohammadzadeh, Mahoney, Shao, Keutzer,
  and Gholami]{kvquant2024}
Hooper, C., Kim, S., Mohammadzadeh, H., Mahoney, M.~W., Shao, Y.~S., Keutzer,
  K., and Gholami, A.
\newblock {KVQuant}: Towards 10 million context length {LLM} inference with
  {KV} cache quantization.
\newblock In \emph{Advances in Neural Information Processing Systems
  (NeurIPS)}, 2024.

\bibitem[Hu et~al.(2025)Hu, Cheng, Yang, Xu, Yuan, Yu, Xu, Jiang, and
  Zhou]{ostquant2025}
Hu, X., Cheng, Y., Yang, D., Xu, Z., Yuan, Z., Yu, J., Xu, C., Jiang, Z., and
  Zhou, S.
\newblock {OSTQuant}: Refining large language model quantization with
  orthogonal and scaling transformations for better distribution fitting.
\newblock In \emph{International Conference on Learning Representations
  (ICLR)}, 2025.

\bibitem[Lee et~al.(2024{\natexlab{a}})Lee, Jin, Kim, Kim, and Park]{owq2024}
Lee, C., Jin, J., Kim, T., Kim, H., and Park, E.
\newblock {OWQ}: Outlier-aware weight quantization for efficient fine-tuning
  and inference of large language models.
\newblock In \emph{Proceedings of the AAAI Conference on Artificial
  Intelligence}, volume~38, pp.\  13355--13364, 2024{\natexlab{a}}.

\bibitem[Lee et~al.(2024{\natexlab{b}})Lee, Lee, and Sim]{tender2024}
Lee, J., Lee, W., and Sim, J.
\newblock Tender: Accelerating large language models via tensor decomposition
  and runtime requantization.
\newblock In \emph{Proceedings of the 51st Annual International Symposium on
  Computer Architecture (ISCA)}, pp.\  1048--1062, 2024{\natexlab{b}}.

\bibitem[Li et~al.(2025)Li, Lin, Zhang, Cai, Li, Guo, Xie, Meng, Zhu, and
  Han]{svdquant2025}
Li, M., Lin, Y., Zhang, Z., Cai, T., Li, X., Guo, J., Xie, E., Meng, C., Zhu,
  J.-Y., and Han, S.
\newblock {SVDQuant}: Absorbing outliers by low-rank components for 4-bit
  diffusion models.
\newblock In \emph{International Conference on Learning Representations
  (ICLR)}, 2025.

\bibitem[Lin et~al.(2024{\natexlab{a}})Lin, Xu, Wu, Cui, Zhang, Mou, Song, Sun,
  and Wei]{duquant2024}
Lin, H., Xu, H., Wu, Y., Cui, J., Zhang, Y., Mou, L., Song, L., Sun, Z., and
  Wei, Y.
\newblock {DuQuant}: Distributing outliers via dual transformation makes
  stronger quantized {LLMs}.
\newblock In \emph{Advances in Neural Information Processing Systems
  (NeurIPS)}, 2024{\natexlab{a}}.

\bibitem[Lin et~al.(2024{\natexlab{b}})Lin, Tang, Tang, Yang, Chen, Wang, Xiao,
  Dang, Gan, and Han]{awq2024}
Lin, J., Tang, J., Tang, H., Yang, S., Chen, W.-M., Wang, W.-C., Xiao, G.,
  Dang, X., Gan, C., and Han, S.
\newblock {AWQ}: Activation-aware weight quantization for {LLM} compression and
  acceleration.
\newblock In \emph{Conference on Machine Learning and Systems (MLSys)},
  2024{\natexlab{b}}.

\bibitem[Lin et~al.(2025)Lin, Tang, Yang, Zhang, Xiao, Gan, and
  Han]{lin2025qserve}
Lin, Y., Tang, H., Yang, S., Zhang, Z., Xiao, G., Gan, C., and Han, S.
\newblock {QServe}: {W4A8KV4} quantization and system co-design for efficient
  {LLM} serving.
\newblock In \emph{Conference on Machine Learning and Systems (MLSys)}, 2025.

\bibitem[Liu et~al.(2026)Liu, Meng, Luo, Zhao, Zhang, and Ma]{micromix2026}
Liu, W., Meng, H., Luo, Y., Zhao, Y., Zhang, P., and Ma, X.
\newblock {MicroMix}: Efficient mixed-precision quantization with microscaling
  formats for large language models.
\newblock In \emph{International Conference on Learning Representations
  (ICLR)}, 2026.

\bibitem[Liu et~al.(2024{\natexlab{a}})Liu, Oguz, Zhao, Chang, Stock, Mehdad,
  Shi, Krishnamoorthi, and Chandra]{llmqat2024}
Liu, Z., Oguz, B., Zhao, C., Chang, E., Stock, P., Mehdad, Y., Shi, Y.,
  Krishnamoorthi, R., and Chandra, V.
\newblock {LLM-QAT}: Data-free quantization aware training for large language
  models.
\newblock In \emph{Findings of the Association for Computational Linguistics:
  ACL 2024}, pp.\  467--484, 2024{\natexlab{a}}.
\newblock \doi{10.18653/v1/2024.findings-acl.26}.

\bibitem[Liu et~al.(2024{\natexlab{b}})Liu, Yuan, Jin, Zhong, Xu, Braverman,
  Chen, and Hu]{kivi2024}
Liu, Z., Yuan, J., Jin, H., Zhong, S., Xu, Z., Braverman, V., Chen, B., and Hu,
  X.
\newblock {KIVI}: A tuning-free asymmetric 2bit quantization for {KV} cache.
\newblock In \emph{Proceedings of the 41st International Conference on Machine
  Learning (ICML)}, Proceedings of Machine Learning Research,
  2024{\natexlab{b}}.

\bibitem[Liu et~al.(2025)Liu, Zhao, Fedorov, Soran, Choudhary, Krishnamoorthi,
  Chandra, Tian, and Blankevoort]{liu2025spinquant}
Liu, Z., Zhao, C., Fedorov, I., Soran, B., Choudhary, D., Krishnamoorthi, R.,
  Chandra, V., Tian, Y., and Blankevoort, T.
\newblock {SpinQuant}: {LLM} quantization with learned rotations.
\newblock In \emph{International Conference on Learning Representations
  (ICLR)}, 2025.

\bibitem[Merity et~al.(2017)Merity, Xiong, Bradbury, and Socher]{wikitext}
Merity, S., Xiong, C., Bradbury, J., and Socher, R.
\newblock Pointer sentinel mixture models.
\newblock In \emph{International Conference on Learning Representations
  (ICLR)}, 2017.

\bibitem[Saxena et~al.(2025)Saxena, Sharify, Roy, and Wang]{saxena2025resq}
Saxena, U., Sharify, S., Roy, K., and Wang, X.
\newblock {ResQ}: Mixed-precision quantization of large language models with
  low-rank residuals.
\newblock In \emph{Proceedings of the 42nd International Conference on Machine
  Learning (ICML)}, volume 267 of \emph{Proceedings of Machine Learning
  Research}, pp.\  53095--53114, 2025.

\bibitem[Shao et~al.(2024)Shao, Chen, Zhang, Xu, Zhao, Li, Zhang, Gao, Qiao,
  and Luo]{omniquant2024}
Shao, W., Chen, M., Zhang, Z., Xu, P., Zhao, L., Li, Z., Zhang, K., Gao, P.,
  Qiao, Y., and Luo, P.
\newblock {OmniQuant}: Omnidirectionally calibrated quantization for large
  language models.
\newblock In \emph{International Conference on Learning Representations
  (ICLR)}, 2024.

\bibitem[Sun et~al.(2025)Sun, Liu, Bai, Bao, Zhao, Li, Hu, Yu, Hou, Yuan,
  Jiang, Liu, and Yao]{flatquant2025}
Sun, Y., Liu, R., Bai, H., Bao, H., Zhao, K., Li, Y., Hu, J., Yu, X., Hou, L.,
  Yuan, C., Jiang, X., Liu, W., and Yao, J.
\newblock {FlatQuant}: Flatness matters for {LLM} quantization.
\newblock In \emph{Proceedings of the 42nd International Conference on Machine
  Learning (ICML)}, Proceedings of Machine Learning Research, 2025.

\bibitem[Tseng et~al.(2024{\natexlab{a}})Tseng, Chee, Sun, Kuleshov, and
  De~Sa]{quipsharp2024}
Tseng, A., Chee, J., Sun, Q., Kuleshov, V., and De~Sa, C.
\newblock {QuIP\#}: Even better {LLM} quantization with {Hadamard} incoherence
  and lattice codebooks.
\newblock In \emph{Proceedings of the 41st International Conference on Machine
  Learning (ICML)}, 2024{\natexlab{a}}.

\bibitem[Tseng et~al.(2024{\natexlab{b}})Tseng, Sun, Hou, and De~Sa]{qtip2024}
Tseng, A., Sun, Q., Hou, D., and De~Sa, C.
\newblock {QTIP}: Quantization with trellises and incoherence processing.
\newblock In \emph{Advances in Neural Information Processing Systems
  (NeurIPS)}, 2024{\natexlab{b}}.

\bibitem[Xiao et~al.(2023)Xiao, Lin, Seznec, Wu, Demouth, and
  Han]{xiao2023smoothquant}
Xiao, G., Lin, J., Seznec, M., Wu, H., Demouth, J., and Han, S.
\newblock {SmoothQuant}: Accurate and efficient post-training quantization for
  large language models.
\newblock In \emph{Proceedings of the 40th International Conference on Machine
  Learning (ICML)}, volume 202 of \emph{Proceedings of Machine Learning
  Research}, 2023.

\bibitem[Yamamoto et~al.(2015)Yamamoto, Nakatsukasa, Yanagisawa, and
  Fukaya]{cholqr}
Yamamoto, Y., Nakatsukasa, Y., Yanagisawa, Y., and Fukaya, T.
\newblock Roundoff error analysis of the {CholeskyQR2} algorithm.
\newblock \emph{Electronic Transactions on Numerical Analysis}, 44:\penalty0
  306--326, 2015.

\bibitem[Yang et~al.(2024)Yang, Yang, Zhang, Hui, Zheng, Yu, Li, Liu, Huang,
  Wei, Lin, Yang, Tu, Zhang, Yang, Yang, Zhou, Lin, Dang, Lu, Bao, Yang, Yu,
  Li, Xue, Zhang, Zhu, Men, Lin, Li, Tang, Xia, Ren, Ren, Fan, Su, Zhang, Wan,
  Liu, Cui, Zhang, and Qiu]{qwen25}
Yang, A., Yang, B., Zhang, B., Hui, B., Zheng, B., Yu, B., Li, C., Liu, D.,
  Huang, F., Wei, H., Lin, H., Yang, J., Tu, J., Zhang, J., Yang, J., Yang, J.,
  Zhou, J., Lin, J., Dang, K., Lu, K., Bao, K., Yang, K., Yu, L., Li, M., Xue,
  M., Zhang, P., Zhu, Q., Men, R., Lin, R., Li, T., Tang, T., Xia, T., Ren, X.,
  Ren, X., Fan, Y., Su, Y., Zhang, Y., Wan, Y., Liu, Y., Cui, Z., Zhang, Z.,
  and Qiu, Z.
\newblock Qwen2.5 technical report.
\newblock \emph{arXiv preprint arXiv:2412.15115}, 2024.

\bibitem[Yao et~al.(2024)Yao, Wu, Li, Youn, and He]{zeroquantv2}
Yao, Z., Wu, X., Li, C., Youn, S., and He, Y.
\newblock Exploring post-training quantization in {LLMs} from comprehensive
  study to low rank compensation.
\newblock In \emph{Proceedings of the AAAI Conference on Artificial
  Intelligence}, volume~38, pp.\  19377--19385, 2024.
\newblock \doi{10.1609/aaai.v38i17.29908}.

\bibitem[Yi et~al.(2024)Yi, Liu, Zhang, Li, Zhang, Lin, and Zhou]{rrs2024}
Yi, K., Liu, Z., Zhang, J., Li, C., Zhang, T., Lin, J., and Zhou, J.
\newblock Rotated runtime smooth: Training-free activation smoother for
  accurate {INT4} inference.
\newblock \emph{arXiv preprint arXiv:2409.20361}, 2024.

\bibitem[Zhang et~al.(2024)Zhang, Cheng, Constantinides, and Zhao]{lqer2024}
Zhang, C., Cheng, J., Constantinides, G.~A., and Zhao, Y.
\newblock {LQER}: Low-rank quantization error reconstruction for {LLMs}.
\newblock In \emph{Proceedings of the 41st International Conference on Machine
  Learning (ICML)}, Proceedings of Machine Learning Research, 2024.

\bibitem[Zhao et~al.(2024)Zhao, Lin, Zhu, Ye, Chen, Zheng, Ceze, Krishnamurthy,
  Chen, and Kasikci]{zhao2024atom}
Zhao, Y., Lin, C.-Y., Zhu, K., Ye, Z., Chen, L., Zheng, S., Ceze, L.,
  Krishnamurthy, A., Chen, T., and Kasikci, B.
\newblock Atom: Low-bit quantization for efficient and accurate {LLM} serving.
\newblock In \emph{Conference on Machine Learning and Systems (MLSys)}, 2024.

\bibitem[Zheng et~al.(2026)Zheng, Song, and Liu]{mixllm2026}
Zheng, Z., Song, X., and Liu, C.
\newblock {MixLLM}: {LLM} quantization with global mixed-precision between
  output-features and highly-efficient system design.
\newblock In \emph{Conference on Machine Learning and Systems (MLSys)}, 2026.

\end{thebibliography}

\section*{AI Usage Disclosure}

The research ideas, the design of the method, the experiment plan and the
decisions on what to keep or discard are the authors'. We used Claude Code
(Anthropic's Claude models) as follows.

\emph{Ideation.} The tool helped us analyze the impact of each idea and work out details, so that we could choose the direction.

\emph{Literature search.} It helped us find related work and published baseline numbers.

\emph{Code and experiments.} We designed the implementation at a high level and specified the experiments. Under our direction, the tool implemented the changes to the open-source baseline pipeline, the GPU kernels, the latency harness, and figure scripts. It launched and monitored the runs and collected their results, which we reviewed.

\emph{Data.} No data or results were generated by AI. Every number comes from a measured run or is quoted from the cited paper.

\emph{Writing.} The authors set the narrative and wrote and revised the paper. The tool drafted some passages from our instructions and polished the writing.

The authors reviewed all of this work and take full responsibility for the
paper.

\clearpage
\onecolumn
\appendix
\makeatletter
\setlength{\@fptop}{0pt}\setlength{\@fpsep}{12pt plus 2pt}\setlength{\@fpbot}{0pt plus 1fil}
\makeatother
\renewcommand{\floatpagefraction}{0.8}

\section{Bit accounting}
\label{app:bits}

Per token, a layer input of width $H$ with $r$ high-precision directions costs
$8r + 4(H-r) = 4H + 4r$ bits in the offline method. \method{} with a rank-$k$ component costs
\begin{equation*}
\underbrace{8(r-k)}_{\text{calibrated}} + \underbrace{4(H-r+k)}_{\text{residual}}
 + \underbrace{4k}_{\text{coefficients}} = 4H + 4r ,
\end{equation*}
because the residual is $H-r+k$ wide: removing the low-rank component
subtracts energy, not coordinates. With 8-bit coefficients the cost would be
$4H + 4r + 4k$; equal bits would then require $r-2k$ calibrated directions.
The down-projection input has $r = 0$ in the offline method, so its $k$
coefficients add $4k$ bits per token. For $k=32$ that is $+0.21\%$ of all
activation bits on Llama-3.2-1B, $+0.11\%$ on Llama-3.1-8B and $+0.06\%$ on
Llama-3.1-70B, and the share grows linearly with $k$. Per-token scales and zero points are
the same in both methods except for one extra scale for the coefficients.

\paragraph{Larger budgets.} Without the substitution, a rank-$k$ component on
top of the offline method's $r$ costs $8r + 4(H-r) + 4k = 4H + 4(r+k)$ bits per
token, exactly the cost of the offline method with $r + k$ calibrated
directions. The comparison in \cref{tab:bits} is therefore at equal bits
at every budget, except for the down-projection input, where the offline
method has no calibrated block to enlarge and \method{}'s $k$ coefficients are
additional.

\section{Exactness of the reassociated layer}
\label{app:exact}

For an orthonormal $V \in \mathbb{R}^{k\times c}$ and $C = X_c V^\top$,
$X_c = (X_c - CV) + CV$, so $X_c W_c^\top = R\,W_c^\top + C\,(V W_c^\top)$.
\Cref{eq:method} quantizes the two terms separately; without quantization it
reproduces the layer exactly, for any $V$. In our implementation the
coefficients are quantized first and $R$ is computed from the quantized
coefficients, so the coefficients' rounding error is carried by the residual
and quantized there rather than lost. $V W_c^\top$ is formed from the
already-quantized weight and kept at 8 bits.

\section{Bases and their evaluation}
\label{app:fit}

\paragraph{Calibrated basis.} For each quantized input, the evaluation runs 16
WikiText-2 training sequences through the already-quantized preceding layers,
accumulates the second moment of the input's low-precision coordinates (after
the online Hadamard transform at the down-projection input), and keeps its
top-$k$ eigenvectors. The basis is fixed before any test sequence is seen.

\paragraph{Per-prompt and shrunk bases.} The per-prompt basis is estimated on
the fit sample (64-token blocks every 256 tokens) with a randomized range
finder~\cite{halko2011finding}: a Gaussian sketch, one power iteration, and one
CholeskyQR. The shrunk basis (\cref{eq:shrink}) uses a rank-512 factor of the
calibrated second moment.

\paragraph{Held-out scoring.} A per-prompt basis quantizes the tokens it was
fitted on, so their perplexity reflects an in-sample fit. Held-out perplexity
scores only tokens outside the fit sample and after the first quarter of the
prompt, for which no basis has seen the token.

\section{Motivation measurements}
\label{app:motivation}

\Cref{tab:motivation} gives the measurements of \cref{sec:motivation} per
layer input, averaged over the 16 layers of Llama-3.2-1B and 16 held-out
prompts of 2048 tokens. The calibrated subspace is the top-$H/8$ eigenspace
of the input's second moment over 64 WikiText-2 training sequences; the
per-prompt directions are the exact top-32 right singular vectors of the first
quarter of each prompt, in the complement of the first $r - 32$ calibrated
directions. The 4-bit error is that of per-token asymmetric quantization
after a random rotation within the low-precision part, divided by the norm of
the full activation.

\begin{table}[!ht]
\centering\small
\caption{Motivation measurements per layer input, Llama-3.2-1B ($r = H/8$,
$k = 32$).}
\label{tab:motivation}
\begin{tabular}{lrrrr}
\toprule
 & attn.\ in & attn.\ out & FFN in & down-proj.\ in \\
\midrule
Energy in calibrated top-$r$, calibration set & 76.0\% & 83.3\% & 62.3\% & 74.0\% \\
\quad held-out prompts                        & 74.9\% & 81.1\% & 59.7\% & 70.6\% \\
Prompt's own top-$r$                          & 88.5\% & 94.0\% & 82.8\% & 96.5\% \\
$r-k$ calibrated $+$ $k$ per prompt           & 78.3\% & 84.5\% & 65.8\% & 74.5\% \\
4-bit error, calibrated only                  & 0.066 & 0.057 & 0.083 & 0.076 \\
\quad with the per-prompt directions removed  & 0.061 & 0.052 & 0.077 & 0.071 \\
Overlap of two prompts' per-prompt subspaces & 4.0\% & 5.6\% & 3.4\% & 1.1\% \\
\quad independent random subspaces            & 1.8\% & 1.8\% & 1.8\% & 0.4\% \\
\bottomrule
\end{tabular}
\end{table}

\section{Training the rotation}
\label{app:rotation}

Rotations trained with quantization in the loop (100 Cayley steps, W16A4KV4)
improve the offline method slightly on Llama-3.2-1B (12.005 $\to$ 11.960 at 2k,
10.561 $\to$ 10.514 at 8k). For \method{} with a per-prompt basis, a rotation
trained with the low-rank component in the loop beats one trained without it (11.608 vs 11.652), but
neither beats a random rotation (11.595). We therefore use random rotations for
both methods.

\section{Additional results}
\label{app:more}

\begin{table}[!ht]
\centering\small
\caption{Low-rank component on subsets of the layer inputs: accuracy
(Llama-3.2-1B, 8k, WikiText-2; ResQ 10.561) and overhead (8B shapes, 8k,
batch 1, RTX~4090, median of three; overhead over ResQ). Gain: share of the
full method's improvement. The down-projection fit runs inside the SwiGLU
kernel only when the FFN input carries the component, which is why removing
the FFN input raises the prefill overhead.}
\label{tab:perinput}
\begin{tabular}{lrrrr}
\toprule
Component at & ppl & gain & prefill & decode \\
\midrule
All four inputs & \textbf{10.032} & 100\% & +4.2\% & +7.8\% \\
All but attention input  & 10.060 & 95\% & +2.9\% & +6.9\% \\
All but attention output & 10.080 & 91\% & +3.2\% & +5.8\% \\
All but FFN input        & 10.115 & 84\% & +5.5\% & +5.4\% \\
FFN input and down-proj. & 10.114 & 84\% & +1.8\% & +5.0\% \\
Down-projection only     & 10.179 & 72\% & +2.7\% & +2.3\% \\
All but down-projection  & 10.571 & $<0$ & +1.1\% & +6.1\% \\
\bottomrule
\end{tabular}
\end{table}

\begin{table}[!ht]
\centering\small
\caption{The component's basis on every model, WikiText-2, equal bits, final
setting: per prompt (\method{}), calibrated offline, and shrunk toward the
calibrated basis (\cref{eq:shrink}). Full-sequence perplexity; in parentheses,
perplexity on held-out tokens (\cref{sec:eval}). Bold: lowest in each row.}
\label{tab:basismodels}
\setlength{\tabcolsep}{5pt}
\begin{tabular}{llrrrr}
\toprule
Model & Ctx & ResQ & Per prompt & Calibrated & Shrunk \\
\midrule
Llama-3.2-1B & 2k  & 12.005 & \textbf{11.329} & 11.485 & 11.353 \\
 & 8k  & 10.561 (10.100) & \textbf{10.032} (\textbf{9.628}) & 10.112 (9.684) & 10.068 (9.679) \\
 & 16k & 10.210 (9.756) & \textbf{9.754} (\textbf{9.338}) & 9.802 (9.391) & 9.785 (9.404) \\
\midrule
Llama-3.2-3B & 2k  & 8.680 & \textbf{8.451} & 8.552 & 8.474 \\
 & 8k  & 7.731 (7.431) & \textbf{7.552} (\textbf{7.262}) & 7.607 (7.315) & 7.567 (7.283) \\
\midrule
Llama-3.1-8B & 2k  & 7.093 & \textbf{6.835} & 6.913 & 6.861 \\
 & 8k  & 6.335 (6.127) & \textbf{6.165} (\textbf{5.962}) & 6.204 (5.992) & 6.172 (5.973) \\
\midrule
Qwen2.5-7B & 2k  & 8.396 & 8.354 & 8.467 & \textbf{8.283} \\
 & 8k  & 7.679 (7.500) & 7.566 (7.384) & \textbf{7.477} (\textbf{7.293}) & 7.552 (7.381) \\
\midrule
Qwen2.5-32B & 2k  & 5.564 & \textbf{5.431} & 5.498 & 5.451 \\
 & 8k  & 4.945 (4.759) & \textbf{4.842} (\textbf{4.665}) & 4.889 (4.704) & 4.867 (4.695) \\
\midrule
Llama-3.1-70B & 2k  & 4.024 & \textbf{3.758} & 3.878 & 3.793 \\
 & 8k  & 3.648 (3.542) & \textbf{3.426} (\textbf{3.350}) & 3.501 (3.408) & 3.451 (3.380) \\
\bottomrule
\end{tabular}
\end{table}

\paragraph{Spending more bits.} A rank-$k$ component with 4-bit coefficients
costs exactly as much as $k$ more protected directions (\cref{app:bits}). At
every budget we measured, spending the bits on the component is better than
enlarging the protected block (\cref{tab:bits}).

\begin{table}[!ht]
\centering\small
\caption{More bits, Llama-3.2-1B, 8k: ResQ with a larger protected block
against its 256 plus \method{}'s low-rank component of rank 64, 128 or 256;
at 256, \method{} as in the main results (224 + 32).}
\label{tab:bits}
\begin{tabular}{lrrr}
\toprule
Budget & ResQ & \method{} & \\
\midrule
256 & 10.561 & \textbf{10.032} & (224 + 32) \\
320 & 10.407 & \textbf{9.952}  & (256 + 64) \\
384 & 10.242 & \textbf{9.843}  & (256 + 128) \\
512 & 10.106 & \textbf{9.731}  & (256 + 256) \\
\bottomrule
\end{tabular}
\end{table}

\FloatBarrier
\section{Full results}
\label{app:results}

\begin{table}[!h]
\centering\small
\caption{Zero-shot accuracy per task (\%), first 500 examples per task; MMLU:
first 25 questions per subject. $^\ast$ Reported by~\citet{saxena2025resq}
under their protocol, not comparable to the other rows: full test sets,
normalized accuracy on ARC, HellaSwag, OBQA and PIQA, 0-shot MMLU; Llama-3 for
the 8B and 70B rows (QuIK at 70B does not fit in memory in our runs). Bold: best
of our own runs for each model.}
\label{tab:zs_full}
\begin{tabular}{llrrrrrrrrrr}
\toprule
Model & Method & ARC-c & ARC-e & BoolQ & HellaS. & OBQA & PIQA & SIQA & WinoG. & Mean & MMLU \\
\midrule
Llama-3.2-1B  & QuIK    & 27.8 & 52.2 & 59.0 & 40.8 & 20.0 & 66.0 & 37.4 & 54.6 & 44.7 & 28.0 \\
              & SpinQuant$^\ast$ & 32.3 & 51.8 & 59.3 & 55.4 & 30.4 & 67.7 & 38.6 & 54.7 & 48.8 & 25.6 \\
              & QuaRot  & 25.8 & 57.2 & 57.4 & 42.8 & 19.6 & 65.6 & 39.2 & 53.6 & 45.1 & 27.7 \\
              & ResQ    & 28.4 & 58.2 & \textbf{59.6} & \textbf{44.0} & 23.6 & 72.0 & 40.6 & 56.6 & 47.9 & 28.1 \\
              & \method{} & \textbf{29.6} & \textbf{59.6} & 55.6 & 43.4 & \textbf{25.6} & \textbf{72.2} & \textbf{41.4} & \textbf{58.0} & \textbf{48.2} & \textbf{33.7} \\
\midrule
Llama-3.2-3B  & QuIK    & 34.0 & 66.2 & 66.0 & 45.6 & 27.4 & 72.4 & 43.2 & 59.4 & 51.8 & 39.1 \\
              & SpinQuant$^\ast$ & 38.9 & 64.8 & 68.0 & 69.1 & 39.4 & 74.9 & 45.1 & 62.9 & 57.9 & 44.2 \\
              & QuaRot  & 37.6 & 66.2 & 71.2 & 47.6 & 27.0 & 75.4 & 43.6 & 62.6 & 53.9 & 43.4 \\
              & ResQ    & 39.0 & 70.2 & 70.8 & 48.4 & 30.8 & \textbf{77.6} & 46.0 & 65.4 & 56.0 & 49.6 \\
              & \method{} & \textbf{40.0} & \textbf{73.0} & \textbf{71.6} & \textbf{49.2} & \textbf{31.0} & 76.8 & \textbf{46.6} & \textbf{66.4} & \textbf{56.8} & \textbf{52.2} \\
\midrule
Llama-3.1-8B  & QuIK    & 41.6 & 75.2 & 76.6 & 48.8 & 30.6 & 75.4 & 42.0 & 71.0 & 57.7 & 54.7 \\
              & SpinQuant$^\ast$ & 48.0 & 75.4 & 75.8 & 75.4 & 43.8 & 77.5 & 45.0 & 69.2 & 63.8 & 56.2 \\
              & QuaRot  & 43.4 & 75.6 & 77.4 & 49.8 & 32.6 & 78.8 & 44.0 & 72.4 & 59.2 & 54.8 \\
              & ResQ    & 45.8 & 77.8 & \textbf{81.6} & 51.8 & \textbf{33.4} & 79.2 & 43.0 & \textbf{73.0} & 60.7 & 58.8 \\
              & \method{} & \textbf{49.8} & \textbf{79.0} & 79.2 & \textbf{52.0} & 32.6 & \textbf{79.4} & \textbf{45.0} & 71.8 & \textbf{61.1} & \textbf{61.9} \\
\midrule
Qwen2.5-7B    & QuIK    & 20.8 & 21.6 & 37.6 & 26.8 & 13.6 & 51.4 & 34.0 & 46.2 & 31.5 & 23.3 \\
              & QuaRot  & 19.8 & 42.0 & 40.4 & 29.8 & 17.8 & 56.4 & 34.2 & 48.8 & 36.1 & 24.2 \\
              & ResQ    & 45.0 & 75.2 & \textbf{80.6} & 53.2 & 31.6 & 80.0 & 51.4 & 68.0 & 60.6 & 69.8 \\
              & \method{} & \textbf{46.6} & \textbf{78.4} & \textbf{80.6} & \textbf{53.4} & \textbf{32.4} & \textbf{80.2} & \textbf{54.2} & \textbf{70.2} & \textbf{62.0} & \textbf{71.1} \\
\midrule
Qwen2.5-32B   & QuIK    & 44.0 & 76.2 & 79.0 & 53.2 & 33.8 & 78.2 & 50.8 & 72.6 & 61.0 & 74.1 \\
              & QuaRot  & 45.8 & 76.0 & 82.8 & 55.8 & 32.0 & 81.4 & 51.4 & 72.6 & 62.2 & 78.9 \\
              & ResQ    & 50.0 & \textbf{78.8} & \textbf{84.8} & 56.8 & 33.2 & 81.4 & 53.6 & 75.4 & 64.3 & 80.0 \\
              & \method{} & \textbf{51.0} & 78.6 & \textbf{84.8} & \textbf{58.4} & \textbf{34.8} & \textbf{82.0} & \textbf{54.2} & \textbf{76.2} & \textbf{65.0} & \textbf{81.5} \\
\midrule
Llama-3.1-70B & QuIK$^\ast$ & 44.5 & 68.9 & 60.7 & 75.0 & 36.4 & 76.1 & 43.2 & 60.4 & 58.2 & 51.1 \\
              & SpinQuant$^\ast$ & 52.0 & 77.3 & 81.7 & 75.6 & 43.8 & 78.8 & 43.4 & 72.8 & 65.7 & 59.4 \\
              & QuaRot  & 51.2 & 81.2 & 81.0 & 53.8 & 33.6 & 80.6 & 46.6 & 77.6 & 63.2 & 71.0 \\
              & ResQ    & 56.6 & 84.0 & 83.0 & 55.4 & \textbf{36.0} & 82.0 & 48.6 & 78.4 & 65.5 & 74.9 \\
              & \method{} & \textbf{59.2} & \textbf{88.2} & \textbf{85.8} & \textbf{57.4} & 35.8 & \textbf{83.2} & \textbf{49.6} & \textbf{79.4} & \textbf{67.3} & \textbf{75.4} \\
\bottomrule
\end{tabular}
\end{table}

\FloatBarrier
\section{Overhead on an A100}
\label{app:a100}

\Cref{tab:overhead_a100} repeats the overhead measurement of
\cref{tab:overhead} on an A100 (80\,GB PCIe), with the shared kernels tuned for
this GPU for both methods. One setting differs: on the A100 the producer
fusions of the fit (\cref{sec:system:fit}) do not pay off at 8k, so it uses
separate kernels at both lengths (with them, the 8k overhead is 10.1\%). The
overhead is larger than on the RTX~4090. The work \method{} adds consists of
thin products, reductions and short synchronized phases, whose speed follows
the GPU's clock and per-SM throughput, where the A100 is behind; the GEMMs
that dominate ResQ's time are bound by memory bandwidth, where it is ahead.
Against FP16 both W4A4 methods gain less than on the RTX~4090, since the
A100's FP16 tensor-core rate is about twice the RTX~4090's while its integer
rates are similar.

\begin{table}[!h]
\centering\small
\caption{\method{} on an A100, Llama-3.1-8B shapes, batch 1, median of three
runs. Top: overhead over ResQ, as in \cref{tab:overhead}. Bottom: speedup
over FP16 (\cref{tab:fp16}), ResQ / \method{}.}
\label{tab:overhead_a100}
\begin{tabular}{lrr}
\toprule
 & 2k & 8k \\
\midrule
Prefill & +20.8\% & +7.2\% \\
Prefill, fit on 1/8 of the prompt & +19.2\% & +4.7\% \\
Decode, per step & +19.2\% & +18.4\% \\
\midrule
Prefill vs FP16 & 1.89 / 1.56$\times$ & 1.71 / 1.59$\times$ \\
Decode vs FP16 & 1.56 / 1.30$\times$ & 1.57 / 1.33$\times$ \\
\bottomrule
\end{tabular}
\end{table}

\end{document}